\documentclass[10pt,journal,compsoc]{IEEEtran}

\ifCLASSOPTIONcompsoc
\usepackage[nocompress]{cite}
\else
\usepackage{cite}
\fi
\ifCLASSINFOpdf
\else
\fi
\usepackage{graphicx}
\usepackage{subfigure} 
\usepackage{amsmath}
\usepackage{amsthm}
\usepackage{amssymb}
\usepackage{algorithm}
\usepackage{algorithmic}
\usepackage{url}
\usepackage{float}
\usepackage{multirow}
\usepackage{color}
\usepackage{soul}

\newtheorem{theorem}{Theorem}[section]
\newtheorem{lemma}[theorem]{Lemma}
\newtheorem{prop}[theorem]{Proposition}
\newtheorem{corollary}[theorem]{Corollary}

\newtheorem{definition}{Definition}

\def \R{\mathbb R}

\def\l{\lambda}

\def\etal{et al.\;}

\newcommand{\X}{\mathbf{X}}

\newcommand{\NM}[2]{\| #1 \|_{#2} }  

\newcommand{\Prox}[2]{\text{prox}_{#1}(#2)}
\newcommand{\SO}[1]{\mathcal{P}_{\mathbf{\Omega}}(#1)}
\newcommand{\Tr}[1]{\text{tr}( #1 ) }
\newcommand{\Span}[1]{\text{span}(#1)}
\newcommand{\QR}[1]{\text{\sf QR}( #1 )}

\newcommand{\f}[1]{f(#1)}
\newcommand{\Diag}[1]{\text{Diag}(#1)}

\usepackage{array}
\newcolumntype{L}[1]{>{\raggedright\let\newline\\\arraybackslash\hspace{0pt}}m{#1}}
\newcolumntype{C}[1]{>{\centering\let\newline  \\\arraybackslash\hspace{0pt}}m{#1}}
\newcolumntype{R}[1]{>{\raggedleft\let\newline \\\arraybackslash\hspace{0pt}}m{#1}}

\begin{document}
	
\markboth{IEEE TRANSACTIONS ON PATTERN ANALYSIS AND MACHINE INTELLIGENCE (Accepted, 2018)}%
{}

	
%
\title{Large-Scale Low-Rank Matrix Learning with Nonconvex Regularizers}



\author{
Quanming Yao, ~\IEEEmembership{Member IEEE},
James T. Kwok, ~\IEEEmembership{Fellow IEEE},
Taifeng Wang, ~\IEEEmembership{Member IEEE},
\\
and Tie-Yan Liu,
~\IEEEmembership{Fellow IEEE}
\IEEEcompsocitemizethanks{
\IEEEcompsocthanksitem Q. Yao is with 4Paradigm Inc, Beijing, China. 
	E-mail: yaoquanming@4Paradigm.com
\IEEEcompsocthanksitem J. Kwok is 
	with the Department of Computer Science and Engineering, 
	Hong Kong University of Science and Technology, Clear Water Bay, 
	Hong Kong.
	E-mail: \{qyaoaa, jamesk\}@cse.ust.hk
\IEEEcompsocthanksitem T. Wang and T. Liu are 
	with Microsoft Research Asia, Beijing, China 100010.
	E-mails: \{taifengw, tyliu\}@microsoft.com
}
}


\IEEEtitleabstractindextext{%
\begin{abstract}
Low-rank modeling has many important applications in computer vision and machine learning. 
While the matrix rank is often approximated by the convex nuclear norm, 
the use of nonconvex low-rank regularizers has demonstrated better empirical performance. 
However, the resulting optimization problem is much more challenging. 
Recent state-of-the-art requires an expensive full SVD in each iteration. 
In this paper, we show that for many commonly-used nonconvex low-rank regularizers, 
the singular values obtained from the proximal operator
can be automatically threshold.
This allows the proximal operator to be efficiently approximated by the power method. 
We then develop a fast proximal algorithm and its accelerated variant
with inexact proximal step. 
It can be guaranteed that
the squared distance between consecutive iterates
converges at a  rate of $O(1/T)$,
where $T$ is the number of iterations.
Furthermore,
we show the proposed algorithm can be parallelized, and the resultant algorithm
achieves nearly linear speedup w.r.t. the number of threads. 
Extensive experiments are performed on matrix completion and robust principal component
analysis. Significant speedup 
over the state-of-the-art
is observed. 
\end{abstract}

}

\maketitle

%

\section{Introduction}
\label{sec:intro}

\IEEEPARstart{L}{ow}-rank matrix 
learning 
is a central issue in many machine learning and computer vision problems. For example, 
matrix completion
\cite{candes2009exact},
which is one of the most successful approaches in collaborative filtering,
assumes that the target rating matrix is low-rank. 
Besides 
collaborative filtering,
matrix completion has also been used on tasks such as
video and image processing \cite{hu2013fast,lu2016nonconvex,gu2016weighted}.
Another important use of low-rank matrix learning is robust principal component analysis (RPCA) \cite{candes2011robust}, 
which assumes that the target 
matrix 
is 
low-rank 
and also corrupted by  sparse noise.  RPCA has been popularly used in computer vision
applications such as shadow removal, background modeling
\cite{candes2011robust,sun2013robust,oh2016partial}, and
robust photometric stereo \cite{wu2010robust}.
Besides,
low-rank matrix learning has also been used in 
face recognition \cite{candes2011robust,sun2013robust},
domain adaptation \cite{li2017domain}
and 
subspace clustering \cite{liu2013robust,xiao2016robust,xiao2015falrr}.

However, minimization of the matrix rank 
is NP-hard \cite{candes2009exact}. To alleviate this problem, a common
approach is to use  
a convex surrogate such as 
the nuclear norm (which is the sum of singular values of the matrix).
It is known that the
nuclear norm is the tightest convex lower bound of the rank.
Though the nuclear norm is non-smooth, the resultant optimization problem can be 
solved efficiently using modern tools such as the
proximal algorithm \cite{ji2009accelerated,mazumder2010spectral,quan2015impute},
Frank-Wolfe algorithm \cite{zhang2012accelerated},
and active subspace selection \cite{hsieh2014nuclear}.

Despite the success of the nuclear norm, recently there have been numerous attempts to use
nonconvex surrogates that better
approximate the rank function.
The key idea is 
that the larger, and thus more informative, singular values
should be less penalized.
Example nonconvex low-rank regularizers include
the capped-$\ell_1$ penalty \cite{zhang2010analysis},
log-sum penalty (LSP) \cite{candes2008enhancing},
truncated nuclear norm (TNN) \cite{hu2013fast,oh2016partial},
smoothly clipped absolute deviation (SCAD) \cite{fan2001variable}, and
minimax concave penalty (MCP) \cite{zhang2010nearly}. 
They have been applied to various computer vision tasks,
such as image denoising \cite{gu2016weighted} and background modeling \cite{oh2016partial}.
Empirically, these nonconvex regularizers achieve
better recovery performance than the convex nuclear norm regularizer.
Recently, theoretical results have also been established \cite{gui2016towards}.

However, the resultant nonconvex optimization problem is much more challenging.
Most existing optimization algorithms that work with the nuclear norm cannot be
applied. A general approach that can still be used is the concave-convex procedure \cite{yuille-03}, which
decomposes the nonconvex regularizer into a difference of convex functions
\cite{zhang2010analysis,hu2013fast}.  However, a sequence of relaxed optimization problems have to be solved,
and can be computationally expensive \cite{gongZLHY2013,li2015accelerated}.
A more efficient approach is the recently proposed iteratively re-weighted nuclear norm (\textsf{IRNN}) algorithm
\cite{lu2016nonconvex}. It is based on the observation that existing nonconvex regularizers are 
concave with
non-increasing
super-gradients.
Each
\textsf{IRNN} 
iteration only involves computing the 
super-gradient of the regularizer and a singular value decomposition (SVD).
However, performing SVD on a $m\times n$ matrix 
takes $O(mn^2)$
time
(assuming $m\geq n$), and can be expensive on 
large
matrices.

Recently, the proximal algorithm has 
been used for nonconvex low-rank matrix learning \cite{hu2013fast,lu2016nonconvex,lu2015generalized,oh2016partial}.
However, it requires the full SVD to solve the proximal operator, 
which can be expensive.
In this paper, we observe that for the commonly-used nonconvex low-rank
regularizers 
	\cite{zhang2010analysis, candes2008enhancing, hu2013fast,oh2016partial, fan2001variable,
	zhang2010nearly},
the singular values obtained from the corresponding proximal operator can be automatically  
thresholded.
One then only needs to find the leading singular values/vectors in order to generate the next iterate. 
Moreover, instead of computing the proximal operator on a large matrix, one only needs to
use the matrix projected onto its leading subspace. The matrix size is  
significantly reduced and the proximal operator can be made much more efficient.
Besides, by using the power method \cite{halko2011finding},
a good approximation of this subspace can be efficiently obtained.  

While the proposed procedure can be readily used with the standard proximal algorithm,
its convergence properties are not directly applicable as
the proximal step here is only approximately solved.
In this paper, we will show that inexactness on the proximal step can be controlled.
The resultant algorithm, which will be called ``Fast Nonconvex Low-rank Learning
(\textsf{FaNCL})",
can be shown to have a
convergence rate of $O(1/T)$ (measured by the squared distance between consecutive
iterates).
This can be further speeded up  using acceleration, leading to the 
\textsf{FaNCL-acc} algorithm.

Effectiveness of the proposed
algorithms is demonstrated on two popular low-rank matrix learning applications,
namely matrix completion and robust principal component analysis (RPCA).  For matrix
completion, we show that additional speedup is possible by exploiting the problem's ``sparse
plus low-rank" structure; whereas for RPCA, we extend the proposed algorithm so that it can
handle the two parameter blocks involved in the RPCA formulation.
With the popularity of  multicore shared-memory platforms, we parallelize
the proposed algorithms so as to handle much larger data sets.

Experiments are performed on both synthetic and real-world data sets.
Results show that the proposed nonconvex low-rank matrix learning algorithms can be several orders faster than the state-of-the-art,
and outperform approaches based on matrix factorization and nuclear norm regularization.
Moreover, the parallelized variants achieve almost linear speedup w.r.t. the number of threads.


In summary,
this paper has three main novelties:
(i) 
Inexactness on the proximal step can be controlled;
(ii) 
Use of acceleration for further speedup; (iii) Parallelization for much larger data sets. 
As can be seen
from Table~\ref{tab:comp}, 
the proposed \textsf{FaNCL-acc} 
is the only  parallelizable,
accelerated inexact
proximal 
algorithm 
on nonconvex problems.

\begin{table}[ht]
\centering
\caption{Comparison of the proposed algorithms with existing algorithms.}
\vspace{-10px}
\begin{tabular}{ c | c | c | C{35px} | c }
	\hline
	method                                     & regularizer & acceleration & proximal step & parallel \\ \hline
	\textsf{APG} \cite{schmidt2011convergence} & convex      & yes          & inexact       & no       \\ \hline
	\textsf{GIST} \cite{gongZLHY2013}          & nonconvex   & no           & exact         & no       \\ \hline
	\textsf{GD} \cite{attouch2013convergence}  & nonconvex   & no           & inexact       & no       \\ \hline
	\textsf{nmAPG}  \cite{li2015accelerated}   & nonconvex   & yes          & exact         & no       \\ \hline
	\textsf{IRNN}  \cite{lu2016nonconvex}      & nonconvex   & no           & exact         & no       \\ \hline
	\textsf{GPG}  \cite{lu2015generalized}     & nonconvex   & no           & exact         & no       \\ \hline
	\textsf{FaNCL}                             & nonconvex   & no           & inexact       & yes      \\ \hline
	\textsf{FaNCL-acc}                         & nonconvex   & yes          & inexact       & yes      \\ \hline
\end{tabular}
\label{tab:comp}
\end{table}

\noindent
\textbf{Notation:}
In the sequel, 
 vectors are denoted by lowercase
boldface, matrices by uppercase boldface,
and the transpose by the superscript $(\cdot)^\top$. 
For a 
square matrix $\mathbf{X}$,
$\Tr{\X}$ is its trace.
For a rectangular matrix $\mathbf{X}$,
$\NM{\X}{F} = \sqrt{\Tr{\X^{\top} \X}}$ is its Frobenius norm, 
and $\NM{\X}{*} = 
\sum_i
\sigma_i( \X )$, where $\sigma_i( \X )$
is the $i$th leading singular value of $\X$, 
is the nuclear norm.
Given $\mathbf{x} = [x_i] \in \R^{m}$, 
$\Diag{\mathbf{x}}$ constructs a $m \times m$ diagonal matrix whose $i$th diagonal element is $x_i$.  
$\mathbf{I}$ denotes the identity matrix.
For a differentiable function $f$, we use $\nabla f$ for its gradient.
For a nonsmooth function,
we use $\partial f$ for its subdifferential,
i.e.,
$\partial f(\mathbf{x}) = \left\lbrace \mathbf{s} 
: f(\mathbf{y}) \ge f(\mathbf{x}) + \mathbf{s}^{\top} (\mathbf{y} - \mathbf{x}) \right\rbrace$.


\section{Background}
\label{sec:review}


\subsection{Proximal Algorithm}
\label{sec:prox}

In this paper, 
we consider 
low-rank matrix learning 
problems  of the form
\begin{equation}
\label{eq:problem}
\min_{\X} F(\X) \equiv f(\X) + \lambda r(\X),
\end{equation} 
where $f$ is a smooth loss, $r$ is a nonsmooth
low-rank regularizer, and 
$\lambda$ is a regularization parameter.
We make the following assumptions on $f$.
\begin{itemize}
\item[A1.] $f$ is not necessarily convex, but is differentiable with $\rho$-Lipschitz continuous
gradient, i.e., $\NM{\nabla \f{\X_1} - \nabla \f{\X_2}}{F} \le \rho \NM{\X_1 - \X_2}{F}$.
Without loss of generality, we assume that $\rho \le 1$.

\item[A2.] $f$ is bounded below, i.e., $\inf \f{\X} > -\infty$,
and $\lim_{\NM{\X}{F} \rightarrow \infty} f(\X) = \infty$.
\end{itemize}

In recent years, the proximal algorithm
\cite{parikh2014proximal} has been popularly used for solving
(\ref{eq:problem}).
At iteration $t$, it produces 
\begin{equation} \label{eq:prosplt}
\X_{t+1} = \Prox{\frac{\lambda}{\tau} r}{\X_t - \frac{1}{\tau} \nabla f(\X_t)},
\end{equation} 
where
$\tau > \rho$ is the stepsize, and
\begin{align}
\Prox{\frac{\l}{\tau} r}{\mathbf{Z}}
\equiv \text{\normalfont Arg}\min_{\X} \frac{1}{2} \NM{\X - \mathbf{Z}}{F}^2  + \frac{\lambda}{\tau} r(\X)
\label{eq:proxoper}
\end{align}
is the proximal operator \cite{parikh2014proximal}.
The proximal step in (\ref{eq:prosplt}) can also be rewritten as
$\X_{t+1} = 
\text{\normalfont Arg}\min_{\mathbf{Y}} 
\Tr{\nabla \f{\X_t}^\top (\mathbf{Y} - \X_t)}
+  
\frac{\tau}{2} \NM{\mathbf{Y} - \X_t}{F}^2 
+ 
\l r(\mathbf{Y})$.
When $f$ and $r$ are convex,
the proximal algorithm converges to the optimal solution at a rate of $O(1/T)$,
where $T$ is the number of iterations.
This can be further accelerated
to 
$O(1/T^2)$, by replacing $\X_t$ in (\ref{eq:prosplt}) with a proper linear combination of $\X_t$ and $\X_{t-1}$
\cite{beck2009fast}.
Recently, the accelerated proximal algorithm has been extended to
problems
where $f$ or $r$ are nonconvex \cite{li2015accelerated,ghadimi2016accelerated}.
The state-of-the-art
is the nonmonotone accelerated proximal gradient
(\textsf{nmAPG}) algorithm \cite{li2015accelerated}. 
As the problem is nonconvex, its convergence rate is still open.  
However, empirically it is much faster.

%


\subsection{Nonconvex Low-Rank Regularizers}
\label{sec:lowrankreg}

For the proximal algorithm to be successful, 
the underlying proximal operator 
has to be efficient.  
The following shows that the proximal operator  
of the 
nuclear norm $\NM{\cdot}{*}$
has a closed-form solution.

\begin{prop}[\!\!\! \cite{cai2010singular}] \label{lem:svt} 
$\Prox{\mu \NM{\cdot}{*}}{\X} 
= \mathbf{U} \left( \mathbf{\Sigma} - \mu \mathbf{I}\right)_{+} \mathbf{V}^{\top}$,
where $\mathbf{U} \mathbf{\Sigma} \mathbf{V}^{\top}$ is the SVD of $\X$, and $\mathbf{A}_{+} = [\max (A_{ij}, 0)]$. 
\end{prop}

Popular 
proximal algorithms 
for nuclear norm minimization
include the \textsf{APG} \cite{toh2010accelerated},
\textsf{Soft-Impute} \cite{mazumder2010spectral}
(and its faster variant \textsf{AIS-Impute} \cite{quan2015impute}),
and active subspace selection\cite{hsieh2014nuclear}.
For \textsf{APG},
the ranks of the iterates 
have to be estimated
heuristically. 
For \textsf{Soft-Impute}, \textsf{AIS-Impute} and active subspace selection,
minimization is performed inside
a subspace.
The smaller the span
of this subspace,
the smaller is the rank of the matrix iterate. 
For good performance, these methods
usually 
require
a much higher
rank.

While the (convex) nuclear norm makes low-rank 
optimization easier,
it may not be a good enough approximation of the matrix rank 
\cite{hu2013fast,lu2016nonconvex,lu2015generalized,gu2016weighted,oh2016partial}. As mentioned in
Section~\ref{sec:intro}, a number of nonconvex surrogates have been recently
proposed.  In this paper, we make the following assumption on the low-rank regularizer
$r$ in (\ref{eq:problem}), which is
satisfied by all nonconvex low-rank regularizers in Table~\ref{tab:lwregu}.

\begin{itemize}
	\item[A3.] $r$ is possibly non-smooth and nonconvex, and 
	of the form $r(\X)=\sum_{i = 1}^m \hat{r}(\sigma_i(\X))$, 
where 
$\hat{r}(0) = 0$
and 
$\hat{r}(\alpha)$ is concave and non-decreasing for $\alpha \ge 0$.
\end{itemize}

\begin{table}[ht]
\centering
\caption{$\hat{r}$'s for some popular nonconvex low-rank regularizers. 
For the TNN regularizer, $\theta \in \{1,\dots,n\}$ is the number of leading singular values that are not penalized;
for SCAD, $\theta > 2$;
and for the others, $\theta > 0$.}
\vspace{-10px}
\begin{tabular}{c | c} \hline
	& $\mu\hat{r}(\sigma_i(\X))$
	\\ \hline
	capped-$\ell_1$ \cite{zhang2010analysis} & $\mu\min(\sigma_i(\X), \theta )$  \\ \hline
	LSP \cite{candes2008enhancing} & $\mu\log \left(\frac{1}{\theta} \sigma_i(\X) + 1\right)$  \\ \hline
	TNN \cite{hu2013fast,oh2016partial}
	& $\begin{cases}
	\mu\sigma_i(\X) & \text{if}\; i  > \theta \\
	0         & \text{otherwise}
	\end{cases}$
	\\ \hline
	SCAD \cite{fan2001variable} & 
	$\begin{cases}
	\mu \sigma_i(\X)                                                  &
	\text{if}\; \sigma_i(\X) \le \mu             \\
	\frac{-\sigma_i^2(\X) + 2 \theta \mu \sigma_i(\X) - \mu^2}{2(\theta - 1)} &
	\text{if}\; \mu < \sigma_i(\X) \le \theta\mu \\
	\frac{(\theta + 1) \mu^2}{2}                              & 
	\text{otherwise}
	\end{cases}$
	\\ \hline
	MCP \cite{zhang2010nearly} & 
	$\begin{cases}
	\mu \sigma_i(\X) - \frac{\sigma_i^2(\X)}{2 \theta} &
	\text{if}\; \sigma_i(\X) \le \theta \mu\\
	\frac{\theta \mu^2}{2}             & \text{otherwise}
	\end{cases}$
	\\ \hline
\end{tabular}
\label{tab:lwregu}
\end{table}

Recently,
the iteratively reweighted nuclear norm (\textsf{IRNN}) algorithm \cite{lu2016nonconvex}
has been proposed to handle this nonconvex low-rank matrix optimization problem. 
In each iteration,
it solves a subproblem
in which the original nonconvex regularizer is approximated by
a weighted version of the nuclear norm
$\NM{\X}{\mathbf{w}} = \sum_{i = 1}^m w_i \sigma_i(\X)$ 
and $0 \leq w_1 \le \dots \le w_m$.
The subproblem has a closed-form solution,
but SVD is needed which takes $O(m n^2)$ time.
Other solvers that are designed for specific nonconvex low-rank regularizers include \cite{sun2013robust} (for capped-$\ell_1$),
\cite{hu2013fast,oh2016partial} (for TNN),
and \cite{zhang2010nearly} (for MCP).
All these (including \textsf{IRNN}) perform SVD in each iteration,
which takes $O(m n^2)$ time and are slow.

While the proximal algorithm has mostly been used  on convex problems, 
recently it is also applied to nonconvex problems \cite{sun2013robust,hu2013fast,lu2016nonconvex,lu2015generalized,gu2016weighted,oh2016partial}.
In particular, the generalized proximal gradient (\textsf{GPG}) algorithm \cite{lu2015generalized} is a 
proximal algorithm which can handle all the above nonconvex regularizers.
In particular, 
the proximal operator
can be computed as follows.

\begin{prop} [Generalized singular value thresholding (GSVT) 
\cite{lu2015generalized}] 
\label{pr:proxReduce}
For any $r$ satisfying assumption~A3, 
$\Prox{\mu r}{\mathbf{Z}} = \mathbf{U}\Diag{\mathbf{y}^*} \mathbf{V}^{\top}$, where $\mathbf{U} \mathbf{\Sigma} \mathbf{V}^{\top}$ is SVD of $\mathbf{Z}$, 
and $\mathbf{y}^*= [y^*_i]$ with
\begin{equation} 
\label{eq:proRed}
y_i^* \in \text{\normalfont Arg}
\min_{y_i \ge 0} \frac{1}{2} \left(y_i - \sigma_i(\mathbf{Z})\right)^2 + \mu \hat{r}(y_i).
\end{equation}
\end{prop}

In \cite{lu2015generalized}, problem (\ref{eq:proRed}) is solved by fixed-point iteration.
However,
closed-form solutions 
indeed
exist for regularizers in Table~\ref{tab:lwregu}
\cite{gongZLHY2013}.
Nevertheless,
Proposition~\ref{pr:proxReduce}
still involves a full SVD,
which takes $O(m n^2)$ time.

Finally,
unlike nuclear norm minimization,
iterates generated by
algorithms for adaptive nonconvex regularization
(including 
\textsf{IRNN},
\textsf{GPG} and
the regularizer-specific algorithms in 
\cite{sun2013robust, hu2013fast,oh2016partial})
may not be low-rank.


\section{Proposed Algorithm}
\label{sec:proalg}
 
In this section, we show how the proximal algorithm for
nonconvex low-rank matrix regularization
can be made much faster.
First, 
Section~\ref{sec:auto}
shows that
the GSVT operator in Proposition~\ref{pr:proxReduce}
does not need
all singular values, which
motivates the development of an approximate GSVT in Section~\ref{sec:apprSVT}.
This approximation is used in the inexact proximal step in Section~\ref{sec:pstep},
and the whole proximal algorithm is shown in Section~\ref{sec:complete}.
Convergence is analysed in Section~\ref{sec:convana}.
Finally, Section~\ref{sec:accFaNCL} presents 
further speedup with the use of acceleration.


\subsection{Automatic Thresholding of Singular Values}
\label{sec:auto}

The following 
Proposition 
\footnote{All Proofs are in Appendix~A.}
shows $y_i^*$
in (\ref{eq:proRed}) becomes zero when
$\sigma_i(\mathbf{Z})$ is smaller than a regularizer-specific threshold. 

\begin{prop}
\label{pr:proxSolution} 
There exists a threshold $\gamma > 0$
such that 
$y^*_i = 0$
when $\sigma_i(\mathbf{Z}) \le \gamma$.
\end{prop} 

Thus, solving the proximal operator in (\ref{eq:proxoper}) only needs the leading singular values/vectors of $\mathbf{Z}$.
For the nonconvex regularizers in Table~\ref{tab:lwregu}, the following
Corollary shows that
simple closed-form solutions  of $\gamma$ can be obtained 
by examining the optimality conditions of (\ref{eq:proRed}).

\begin{corollary} \label{cor:threh} 
The $\gamma$   values 
for the following regularizers 
are:
\begin{itemize}
\item 
Capped-$\ell_1$: $\gamma = \min\left(\sqrt{2\theta \mu}, \mu\right)$;

\item LSP: $\gamma=\min\left(\frac{\mu}{\theta},\theta\right)$; 

\item TNN: $\gamma = \max\left(\mu, \sigma_{\theta + 1}(\mathbf{Z})\right)$;

\item SCAD: $\gamma = \mu$;

\item MCP: $\gamma = \sqrt{\theta} \mu$ if $0 < \theta < 1$, and $\mu$ otherwise.
\end{itemize}
\end{corollary} 
Corollary~\ref{cor:threh} can also be extended to the nuclear norm. 
Specifically,
it can be shown that $\gamma= \frac{\lambda}{\tau}$,
and $y_i^*=\max\left( \sigma_i(\X_{\text{gd}}) - \frac{\lambda}{\tau}, 0 \right)$.
However, since our focus is on nonconvex regularizers,  
it will not be pursued in the sequel.


\subsection{Approximate GSVT}
\label{sec:apprSVT}

Proposition~\ref{pr:proxReduce} computes the proximal operator using exact SVD.
Due to automatic thresholding of the singular values
in Section~\ref{sec:auto},
this can be made more efficient by using
	partial SVD.
Moreover,  we will show in this Section that
the proximal operator only needs to be 
computed
on a much smaller matrix.


\subsubsection{Reducing the Size of SVD}

Assume that $\mathbf{Z}$ has $\hat{k}$ singular values that are larger than $\gamma$.
We 
then
only need to perform a rank-$k$ SVD on $\mathbf{Z}$ with $k \ge \hat{k}$.
Let the rank-$\hat{k}$ SVD of $\mathbf{Z}$ be $\mathbf{U}_{\hat{k}} \mathbf{\Sigma}_{\hat{k}} \mathbf{V}_{\hat{k}}^{\top}$.
The following Proposition shows that 
$\Prox{\mu r}{\mathbf{Z}}$
can be obtained from the proximal operator on the smaller matrix
$\mathbf{Q}^{\top} \mathbf{Z}$.
\footnote{We noticed a similar result in \cite{oh2018pami} after the conference version of this paper \cite{yao2015fast} has been accepted.  However, \cite{oh2018pami} only considers the case where $r$ is the nuclear norm regularizer.}

\begin{prop} \label{pr:redGSVT} 
Assume that $\mathbf{Q} \in \R^{m \times k}$, where $k \ge \hat{k}$, is orthogonal and
$\Span{\mathbf{U}_{\hat{k}}} \subseteq \Span{\mathbf{Q}}$.
Then,
$\Prox{\mu r}{\mathbf{Z}} = \mathbf{Q} \, \Prox{\mu r}{\mathbf{Q}^{\top} \mathbf{Z}}$.
\end{prop}


\subsubsection{Obtaining an Approximate GSVT}
\label{sec:apprgsvt}

To obtain such a 
$\mathbf{Q}$,
we will use the 
power method 
(Algorithm~\ref{alg:powermethod}).
It has sound approximation guarantee, good empirical performance
\cite{halko2011finding}, and
has been recently used 
to approximate the SVT in nuclear norm minimization 
\cite{hsieh2014nuclear,quan2015impute}.  
As in \cite{hsieh2014nuclear},
we set the number of power iterations to 3.
Warm-start can be used via matrix $\mathbf{R}$ in
Algorithm~\ref{alg:powermethod}. 
This is particularly useful because of the iterative nature of proximal algorithm.
Obtaining an approximate $\mathbf{Q}$ using Algorithm~\ref{alg:powermethod} takes $O(m n k)$ time.
As in \cite{toh2010accelerated,mazumder2010spectral}, the 
PROPACK package \cite{larsen1998lanczos},
which is based on the Lanczos
algorithm,
can also be used to obtain $\mathbf{Q}$ in $O(m n k)$ time.
However, it finds $\mathbf{Q}$ exactly and
cannot benefit from warm-start. Hence, though it has the same time complexity as 
power method, empirically it is much less efficient \cite{quan2015impute}.

\begin{algorithm}[ht]
\caption{\textsf{Powermethod}$(\mathbf{Z}, \mathbf{R})$.}
\begin{algorithmic}[1]
	\REQUIRE $\mathbf{Z} \in \R^{m \times n}$, $\mathbf{R} \in \R^{n \times k}$
	and the number of power iterations $J=3$.
	\STATE $\mathbf{Y}_1 = \mathbf{Z} \mathbf{R}$;
	\FOR{$j = 1, 2, \dots, J$}
	\STATE $\mathbf{Q}_j = \QR{\mathbf{Y}_j}$;  
	
	// QR decomposition (returning only the $\mathbf{Q}$ matrix)
	\STATE $\mathbf{Y}_{j + 1} = \mathbf{Z} (\mathbf{Z}^{\top} \mathbf{Q}_j)$;
	\ENDFOR
	\RETURN $\mathbf{Q}_{J}$. 
\end{algorithmic}
\label{alg:powermethod}
\end{algorithm}

Algorithm~\ref{alg:apprGSVT} shows steps of the approximate GSVT.
Step~1 uses the power method to efficiently obtain an orthogonal matrix $\mathbf{Q}$ that approximates $\Span{\mathbf{U}_{\hat{k}}}$.
Step~2 
performs
a small SVD.
Though this SVD is still exact,
$\mathbf{Q}^{\top} \mathbf{Z}$ is much smaller than $\mathbf{Z}$ ($k \times n$ vs $m \times n$),
and
$\text{SVD}(\mathbf{Q}^{\top} \mathbf{Z})$
takes only $O(n k^2)$ time.
In step~3,
the singular values $\Sigma_{ii}$'s
are thresholded using Corollary~\ref{cor:threh}.
Steps~4-6 obtains an (approximate) $\Prox{\mu r}{\mathbf{Z}}$
using Proposition~\ref{pr:proxReduce}.
The time complexity for GSVT is reduced from $O(m n^2)$ 
to $O(m n k)$.

\begin{algorithm}[ht]
\caption{Approximate GSVT: \textsf{ApproxGSVT}$(\mathbf{Z}, \mathbf{R}, \mu)$.}
\begin{algorithmic}[1]
	\REQUIRE $\mathbf{Z} \in \R^{m \times n}$ 
	and $\mathbf{R} \in \R^{n \times k}$ for warm-start;
	\STATE $\mathbf{Q} = \text{\sf PowerMethod}(\mathbf{Z}, \mathbf{R})$;
	\STATE $[ \mathbf{U}, \mathbf{\Sigma}, \mathbf{V} ] = \text{\sf SVD}(\mathbf{Q}^{\top} \mathbf{Z})$;
	\STATE $a = $ number of $\Sigma_{ii}$'s that are $>\gamma$
	in Corollary~\ref{cor:threh};
	\STATE $\mathbf{U}_a = a$ leading columns of $\mathbf{U}$;
	\STATE $\mathbf{V}_a = a$ leading columns of $\mathbf{V}$;
	\STATE obtain $y^*_i$ from \eqref{eq:proRed} for all $i = 1, \dots, a$;
	\RETURN  low-rank components of $\tilde{\mathbf{X}}$
	($\mathbf{Q} \mathbf{U}_a$, $\Diag{[ y_1^*,\dots,y_{a}^* ]}$ and $\mathbf{V}_a^{\top}$),
	and $\mathbf{V}$.
\end{algorithmic}
\label{alg:apprGSVT}
\end{algorithm}

\subsection{Inexact Proximal Step}
\label{sec:pstep}

In order to compute the proximal step efficiently,
we will utilize the approximate GSVT in Algorithm~\ref{alg:apprGSVT}.
However,
the resultant proximal step is then inexact.
To ensure convergence of the resultant proximal algorithm,
we need to control the approximation quality of the proximal step.

First,
the following Lemma shows that the objective $F$ is always decreased (as $\tau > \rho$)
when the proximal step is computed exactly.

\begin{lemma}
\label{lem:desc}
{\normalfont  ( $\!\!\!\!$ \cite{gongZLHY2013}, \cite{attouch2013convergence})}
Let $\X_{\text{gd}} = \mathbf{X} - \frac{1}{\tau} \nabla f(\mathbf{X})$.
Then, 
we have
$F( \Prox{\frac{\lambda}{\tau} r}{\X_{\text{gd}}} ) 
\le
F(\mathbf{X}) - \frac{\tau - \rho}{2} 
\NM{\Prox{\frac{\lambda}{\tau} r}{\X_{\text{gd}}} - \mathbf{X}}{F}^2$.
\end{lemma}

Let the approximate 
proximal step 
solution  obtained at the $p$th iteration 
be $\tilde{\X}_p$.
Motivated by 
Lemma~\ref{lem:desc},
we control the quality of $\tilde{\X}_p$ by monitoring the objective value $F$.
Specifically,
we try to ensure that
\begin{align}
F( \tilde{\mathbf{X}}_p )
\le F(\mathbf{X}) - c_1 \NM{\tilde{\mathbf{X}}_p - \mathbf{X}}{F}^2,
\label{eq:decrease}
\end{align} 
where $c_1 = \frac{\tau - \rho }{4}$.
Note that this is less stringent than the condition in Lemma~\ref{lem:desc}.
The procedure is shown in Algorithm~\ref{alg:inexactPS}.  
If \eqref{eq:decrease} holds, 
we accept $\tilde{\mathbf{X}}_p$;
otherwise, 
we improve $\tilde{\mathbf{X}}_p$ by using $\tilde{\mathbf{V}}_{p - 1}$ to warm-start the next iterate.
The following Proposition shows convergence of Algorithm~\ref{alg:inexactPS}.

\begin{prop}
\label{pr:apprGSVT}
If $k \ge \hat{k}_{\X_{\text{gd}}}$,
where $\hat{k}_{\X_{\text{gd}}}$ is the number of singular values in $\X_{\text{gd}}$ larger than $\gamma$,
then
$\lim_{p \rightarrow \infty} \tilde{\mathbf{X}}_p = \Prox{\frac{\lambda}{\tau} r}{\X_{\text{gd}}}$.
\end{prop}

\begin{algorithm}[ht]
\caption{Inexact proximal step: \textsf{InexactPS}$(\mathbf{X}, \mathbf{R})$.}
\begin{algorithmic}[1]
	\REQUIRE $\mathbf{X} \in \R^{m \times n}$, and $\mathbf{R} \in \R^{n \times k}$ for warm-start;
	\STATE $\X_{\text{gd}} = \mathbf{X} - \frac{1}{\tau} \nabla f(\mathbf{X})$;
	\STATE $\tilde{\mathbf{V}}_0 = \mathbf{R}$;
	\FOR{$p = 1,2,\dots$}
	\STATE $[ \tilde{\mathbf{X}}_p, \tilde{\mathbf{V}}_p ]  = $
	\textsf{ApproxGSVT}
	$( \X_{\text{gd}}, \tilde{\mathbf{V}}_{p - 1}, \frac{\lambda}{\tau} )$;
	\STATE \textbf{if} $F( \tilde{\mathbf{X}}_p ) \le F(\mathbf{X}) - c_1 \NM{\tilde{\mathbf{X}}_p - \mathbf{X}}{F}^2$
	\textbf{then} break;
	\ENDFOR
	\RETURN 
	$\tilde{\mathbf{X}}_p$ and $\tilde{\mathbf{V}}_p$.
\end{algorithmic}
\label{alg:inexactPS}
\end{algorithm}

The use of inexact proximal steps has also been considered in \cite{attouch2013convergence,schmidt2011convergence}.  
However, $r$ 
in (\ref{eq:problem})
is assumed to be convex in \cite{schmidt2011convergence}.
Attouch \etal \cite{attouch2013convergence} considered nonconvex $r$, but they
	require a difficult and expensive condition to control inexactness
	(an example is provided in Appendix~B).

\subsection{The Complete Procedure}
\label{sec:complete}

The complete procedure  
for solving (\ref{eq:problem})
is shown in Algorithm~\ref{alg:FaNCL},
and will be called  
FaNCL (\underline{Fa}st \underline{N}on\underline{C}onvex \underline{L}owrank).
Similar to \cite{hsieh2014nuclear,quan2015impute}, we perform warm-start 
using the column spaces of the previous iterates ($\mathbf{V}_{t}$ and $\mathbf{V}_{t - 1}$).
For further speedup, 
we employ a continuation strategy at step~3 as in \cite{mazumder2010spectral,lu2016nonconvex,toh2010accelerated}.
Specifically,
$\lambda_t$ is initialized to a large value
and then
decreases
gradually.

\begin{algorithm}[ht]
\caption{\textsf{FaNCL} \!\! (Fast \! NonConvex Low-rank) \!\! algorithm.}
\begin{algorithmic}[1]
	\REQUIRE choose $\tau > \rho$, $\lambda_0 > \lambda$ and $\nu \in (0,1)$;
	\STATE initialize $\mathbf{V}_0, \mathbf{V}_1 \in \R^{n \times 1}$ as random Gaussian matrices and $\X_1 = 0$;
	\FOR{$t = 1,2,\dots T$}
	\STATE $\lambda_t = (\lambda_{t-1} -\lambda) \nu^t + \lambda$;
	\STATE $\mathbf{R}_t = \QR{[\mathbf{V}_t, \mathbf{V}_{t - 1}]}$; // warm start
	\STATE $[\X_{t + 1}, \mathbf{V}_{t + 1}] =$ \textsf{InexactPS}$(\X_t, \mathbf{R}_t)$;
	\ENDFOR 
	\RETURN $\X_{T + 1}$.
\end{algorithmic}
\label{alg:FaNCL}
\end{algorithm}

Assume that evaluations of $f$ and $\nabla f$ take 
$O(m n)$ time,
which is valid for many applications such as matrix completion and RPCA.
Let $r_t$ be the rank of $\X_t$ at the $t$th iteration, and $k_t = r_t + r_{t - 1}$.  In
Algorithm~\ref{alg:FaNCL}, step~4 takes $O(n k_t^2)$ time; and step~5 takes $O(m n p k_t)$ time
as $\mathbf{R}_t$ has $k_t$ columns.
The iteration time complexity 
is 
thus
$O(m n p k_t)$.
In the experiment, we set $p = 1$,
which is 
enough to guarantee (\ref{eq:decrease})
empirically.
The iteration time complexity 
	of Algorithm~\ref{alg:FaNCL} is thus reduced to
	$O( m n k_t )$.
In contrast, exact GSVT takes $O(m n^2)$ time, and is much slower as $k_t \ll n$.
Besides,
	the space complexity of
	Algorithm~\ref{alg:FaNCL} is $O( m n )$.

\subsection{Convergence Analysis}
\label{sec:convana}

The inexact proximal algorithm is first considered in
\cite{schmidt2011convergence}, which assumes $r$ to be convex.
This does not hold here as the regularizer is nonconvex.
The nonconvex
extension is considered in \cite{attouch2013convergence}.  However, 
it assumes the Kurdyka-Lojasiewicz condition  \cite{bolte2010characterizations} 
on $f$, which
does not hold
for $C^{\infty}$ functions
(including the commonly used square loss)
in general.
On the other hand,
we only assume that $f$ is Lipschitz-smooth.
Besides,
as discussed in Section~\ref{sec:pstep},
they use an 
expensive
condition to control inexactness of the proximal step.
Thus, their analysis cannot 
be applied here.  

In the following, we first
show that $r$,
similar to $\hat{r}$ in Assumption~A3 \cite{gongZLHY2013},
can be decomposed as a difference of convex functions.

\begin{prop} \label{pr:dc}
$r$ can be decomposed as $\breve{r} - \tilde{r}$,
where $\breve{r}$ and $\tilde{r}$ are convex.
\end{prop}

Based on this decomposition, we introduce the definition of critical point.

\begin{definition}
[\!\cite{hiriart85}]
If $\mathbf{0} \in \nabla f(\X) + \lambda \left( \partial \breve{r}(\X) - \partial \tilde{r}(\X) \right) $,
then $\X$ is a {\em critical point} of $F$.
\label{def:crti}
\end{definition}

The following Proposition
shows that Algorithm~\ref{alg:FaNCL} generates a bounded sequence.

\begin{prop} \label{pr:boundseq}
The sequence $\{\X_t\}$ generated from Algorithm~\ref{alg:FaNCL} is bounded,
and has at least one limit point.
\end{prop}

Let $\mathcal{G}_{\frac{\lambda}{\tau} r}(\X_t) = \X_t - \Prox{\frac{\lambda}{\tau}r}{\X_t - \frac{1}{\tau} \nabla f(\X_t)}$,
which is known as the proximal mapping of $F$ at $\X_t$ \cite{parikh2014proximal}.
If $\mathcal{G}_{\frac{\lambda}{\tau} r}(\X_t) = 0$,
$\mathbf{X}_t$ is a critical point of \eqref{eq:problem}
\cite{attouch2013convergence,gongZLHY2013}.
This motivates the use of $\| \mathcal{G}_{\frac{\lambda}{\tau} r}(\X_t) \|_2^2$ to measure convergence 
in \cite{ghadimi2016accelerated}.
However, $\| \mathcal{G}_{\frac{\lambda}{\tau} r}(\X_t) \|_2^2$
cannot be used here as $r$ is nonconvex and the proximal step is inexact.
As 
Proposition~\ref{pr:boundseq}
guarantees
the existence of limit points,
we use $\NM{\X_{t + 1} - \X_t}{F}^2$ instead to measure convergence.
If the proximal step is exact, 
$\| \mathcal{G}_{\frac{\lambda}{\tau} r}(\X_t) \|_F^2 = \| \X_{t + 1} - \X_t \|_F^2$.
The following Corollary
shows convergence of Algorithm~\ref{alg:FaNCL}.

\begin{corollary} \label{cor:rate}
$\min_{t = 1, \dots, T} \NM{\X_{t + 1} - \X_t}{F}^2 \le \frac{F(\X_1) - \inf F}
{c_1 T}$.
\end{corollary}

The following Theorem shows  that
any limit point is also a critical point.

\begin{theorem} \label{the:FaNCL:conv}
Assume that Algorithm~\ref{alg:inexactPS} returns $\mathbf{X}$ only when 
$\mathbf{X} = \Prox{\frac{\lambda}{\tau} r}{\mathbf{X} - \frac{1}{\tau} \nabla
f(\mathbf{X})}$ (i.e.,
the input is 
returned
as output only if  it
is the desired exact proximal step solution).
Let $\{ \X_{t_j} \}$ be
a subsequence of $\{ \X_t \}$ generated by Algorithm~\ref{alg:FaNCL} such that
$\lim_{t_j \rightarrow \infty} \X_{t_j} = \X_*$.
Then, $\X_*$ is a critical point of \eqref{eq:problem}.
\end{theorem}

\subsection{Acceleration}
\label{sec:accFaNCL}

In convex optimization, acceleration has been commonly used to speed up convergence
of proximal algorithms \cite{beck2009fast}.  
Recently, it has also been extended to nonconvex optimization \cite{li2015accelerated,ghadimi2016accelerated}.
A state-of-the-art algorithm is the \textsf{nmAPG} \cite{li2015accelerated}.
In this section, we integrate \textsf{nmAPG} with \textsf{FaNCL}. The whole procedure
is shown in Algorithm~\ref{alg:FaNCLacc}.
The accelerated iterate is obtained in step~4.
If the resultant inexact proximal step solution can achieve a sufficient decrease (step~7)
as in \eqref{eq:decrease},
this iterate is accepted (step~8);
otherwise,
we 
choose the inexact proximal step solution obtained with the non-accelerated iterate
$\mathbf{X}_t$ (step~10).
Note that step~10 
is the same as step~5 of Algorithm~\ref{alg:FaNCL}.
Thus, the iteration time complexity of Algorithm~\ref{alg:FaNCLacc} is at most twice that of Algorithm~\ref{alg:FaNCL}, and still $O(m n k_t)$.  
Besides, its space complexity is $O(m n)$, 
which is the same as Algorithm~\ref{alg:FaNCL}.

There are several major differences between Algorithm~\ref{alg:FaNCLacc} and \textsf{nmAPG}.  First, the proximal step of Algorithm~\ref{alg:FaNCLacc} is only inexact.
To make the algorithm more robust, we do not allow nonmonotonous update (i.e., $F(\X_{t +
1})$ cannot be larger than $F(\X_t)$).
Moreover, we use a simpler acceleration scheme (step~4), in which only $\X_t$ and $\X_{t - 1}$ are involved.  
On matrix completion problems,
this allows using the ``sparse plus low-rank'' structure
\cite{mazumder2010spectral} to greatly reduce the iteration complexity
(Section~\ref{sec:matcomp}).  Finally, we do not require extra comparison of the objective at step~10. This further reduces the iteration complexity.

\begin{algorithm}[ht]
	\caption{Accelerated FaNCL algorithm (\textsf{FaNCL-acc}).}
	\begin{algorithmic}[1]
		\REQUIRE choose $\tau > \rho$, $\lambda_0 > \lambda$, $\delta > 0$ and $\nu \in (0,1)$;
		\STATE initialize $\mathbf{V}_0, \mathbf{V}_1 \in \R^{n}$ as random Gaussian matrices, $\X_0 = \X_1 = 0$
		and $\alpha_0 = \alpha_1 = 1$;
		\FOR{$t = 1,2,\dots T$}
		\STATE $\lambda_t = (\lambda_{t-1} -\lambda) \nu + \lambda$;
		\STATE $\mathbf{Y}_t = \X_t + \frac{\alpha_{t - 1} - 1}{\alpha_t}(\X_t - \X_{t - 1})$;
		\STATE $\mathbf{R}_t = \QR{[\mathbf{V}_t, \mathbf{V}_{t - 1}]}$; // warm start
		\STATE $\X^a_{t + 1} =$ \textsf{InexactPS}$(\mathbf{Y}_t, \mathbf{R}_t)$;
		\IF{$F(\X^a_{t + 1}) \le F(\X_t) - \frac{\delta}{2} \NM{\X^a_{t + 1} - \mathbf{Y}_t}{F}^2$}
		\STATE $\X_{t + 1} = \X^a_{t + 1}$;
		\ELSE
		\STATE $\X_{t + 1} =$ \textsf{InexactPS}$(\X_t, \mathbf{R}_t)$;
		\ENDIF
		\STATE $\alpha_{t + 1} = \frac{1}{2} (\sqrt{4\alpha_t^2 + 1} + 1)$;
		\ENDFOR 
		\RETURN $\X_{T + 1}$.
	\end{algorithmic}
	\label{alg:FaNCLacc}
\end{algorithm}

The following Proposition shows that Algorithm~\ref{alg:FaNCLacc} generates a bounded sequence.

\begin{prop} \label{pr:convacc}
	The sequence $\{\X_t\}$ generated from Algorithm~\ref{alg:FaNCLacc} is bounded,
	and has at least one limit point.
\end{prop}

In Corollary~\ref{cor:rate}, $\NM{\X_{t + 1} - \X_t}{F}^2$ is used to measure progress
before and after the proximal step.  In Algorithm~\ref{alg:FaNCLacc}, the
proximal step may use the accelerated iterate $\mathbf{Y}_t$ or the non-accelerated iterate
$\X_t$. Hence, we
use
$\| \X_{t+1} - \mathbf{C}_t \|_F^2$,
where
$\mathbf{C}_t = \mathbf{Y}_t$ if step~8 is performed, and $\mathbf{C}_t = \X_t$ otherwise.
Similar to Corollary~\ref{cor:rate},
the following shows a $O(1/T)$ convergence rate.

\begin{corollary} \label{cor:rateacc}
$\min_{t = 1, \dots, T} \NM{\X_{t + 1} - \mathbf{C}_t}{F}^2 \le \frac{F(\X_1) - \inf F}{\min(c_1, \delta/2) T}$. 
\end{corollary}

On nonconvex optimization problems, the optimal convergence rate  for first-order methods is
$O(1/T)$ \cite{ghadimi2016accelerated}.  Thus, the convergence rate
of Algorithm~\ref{alg:FaNCLacc} (Corollary~\ref{cor:rateacc}) cannot improve that of
Algorithm~\ref{alg:FaNCL}
(Corollary~\ref{cor:rate}).
However, in practice,
acceleration can still significantly reduce the number of iterations 
on nonconvex problems 
\cite{li2015accelerated,ghadimi2016accelerated}.
On the other hand, as Algorithm~\ref{alg:FaNCLacc} may need a second proximal step 
(step~10), its iteration time complexity can be higher than that of Algorithm~\ref{alg:FaNCL}. 
However, this is much compensated by the speedup in convergence.
As will be demonstrated in Section~\ref{sec:exptmc},
empirically 
Algorithm~\ref{alg:FaNCLacc} is much faster.

The following Theorem shows  that
any limit point of the iterates from
Algorithm~\ref{alg:FaNCLacc} 
is also a critical point.

\begin{theorem} \label{thm:critacc}
Let $\{ \X_{t_j} \}$ be
a subsequence of $\{ \X_t \}$ generated by Algorithm~\ref{alg:FaNCLacc} such that
$\lim_{t_j \rightarrow \infty} \X_{t_j} = \X_*$.
With the assumption in Theorem~\ref{the:FaNCL:conv},
$\X_*$ is a critical point of \eqref{eq:problem}.
\end{theorem}


\begin{table*}[ht]
\centering
\renewcommand{\arraystretch}{1.2}
\caption{Comparison of the iteration time complexities, convergence rates and space
complexity
of various matrix completion solvers.  
Here, $k_t = r_t + r_{t - 1}$, $\nu \in (0,1)$ and integer $T_a > 0$ are constants.  For the
active subspace selection method (\textsf{active}) \cite{hsieh2014nuclear}, $T_s$ is the number of inner iterations  required.}
\vspace{-10px}
\begin{tabular}{c | c | c | c | c}
	\hline
	      regularizer        & method                                                   & convergence rate   & iteration time complexity                                       & space complexity                          \\ \hline
	        (convex)         & \textsf{APG} \cite{ji2009accelerated,toh2010accelerated} & $O(1/T^2)$         & $O(m n r_t)$                                                    & $O(mn)$                                   \\ \cline{2-5}
	      nuclear norm       & \textsf{active} \cite{hsieh2014nuclear}                  & $O(\nu^{T - T_a})$ & $O( \NM{\mathbf{\Omega}}{1} k_t T_{s})$                         & $O((m + n)k_t + \NM{\mathbf{\Omega}}{1})$ \\ \cline{2-5}
	                         & \textsf{AIS-Impute} \cite{quan2015impute}                & $O(1/T^2)$         & $O(\NM{\mathbf{\Omega}}{1} k_t + (m + n)k^2_t)$                 & $O((m + n)k_t + \NM{\mathbf{\Omega}}{1})$ \\ \hline
	fixed-rank factorization & \textsf{LMaFit} \cite{wen2012solving}                    & ---                & $O( \NM{\mathbf{\Omega}}{1} r_t + (m + n)r_t^2)$                & $O((m + n)r_t + \NM{\mathbf{\Omega}}{1})$ \\ \cline{2-5}
	                         & \textsf{ER1MP} \cite{wang2015orthogonal}                 & $O(\nu^T)$         & $O(\NM{\mathbf{\Omega}}{1})$                                    & $O((m + n)r_t + \NM{\mathbf{\Omega}}{1})$ \\ \hline
	       nonconvex         & \textsf{IRNN} \cite{lu2016nonconvex}                     & ---                & $O(m n^2)$                                                      & $O(mn)$                                   \\ \cline{2-5}
	                         & \textsf{GPG} \cite{lu2015generalized}                    & ---                & $O(m n^2)$                                                      & $O(mn)$                                   \\ \cline{2-5}
	                         & \textsf{FaNCL}                                           & $O(1/T)$           & $O\left(  \NM{\mathbf{\Omega}}{1} r_t + (m + n)r_t k_t \right)$ & $O((m + n)r_t + \NM{\mathbf{\Omega}}{1})$ \\ \cline{2-5}
	                         & \textsf{FaNCL-acc}                                       & $O(1/T)$           & $O\left(  \NM{\mathbf{\Omega}}{1} k_t + (m + n)k_t^2 \right)$   & $O((m + n)k_t + \NM{\mathbf{\Omega}}{1})$ \\ \hline
\end{tabular}
\label{tab:timecomp}
\end{table*}


\section{Applications}
\label{sec:applications}

In this section,
we consider two important instances of problem \eqref{eq:problem},
namely, matrix completion \cite{candes2009exact}
and robust principal component analysis (RPCA) \cite{candes2011robust}.
For
matrix completion (Section~\ref{sec:matcomp}),
we will show that the proposed algorithm
can be made even faster and require much less
memory by using the ``sparse plus low-rank" structure of the problem.
In Section~\ref{sec:rpca},
we show how the 
algorithm
can be extended to deal with 
the two parameter blocks
in RPCA.


\subsection{Matrix Completion}
\label{sec:matcomp}

Matrix completion attempts to recover a low-rank matrix $\mathbf{O} \in \R^{m \times
n}$ by observing only some of its elements \cite{candes2009exact}.  
Let the 
observed positions be indicated by  
$\mathbf{\Omega} \in \{0,1\}^{m \times n}$, such that
$\Omega_{ij}=1$ if $O_{ij}$ is observed, and 0 otherwise.
Matrix completion can be formulated as an optimization
problem in (\ref{eq:problem}), with
\begin{align}
F(\X) = \frac{1}{2}\NM{\SO{\X - \mathbf{O}}}{F}^2 + \lambda r(\X),
\label{eq:matcomp}
\end{align}
where $[\SO{\X_{\text{gd}}}]_{ij} = A_{ij}$ if  $\Omega_{ij} = 1$ and $0$ otherwise.

\subsubsection{Utilizing the Problem Structure}

In the following, we focus on 
the accelerated \textsf{FaNCL-acc} algorithm (Algorithm~\ref{alg:FaNCLacc}), 
and show that its time and space complexities 
can be further reduced.
Similar techniques can also be used on 
the simpler non-accelerated \textsf{FaNCL} algorithm (Algorithm~\ref{alg:FaNCL}).

First, consider step~7
(of Algorithm~\ref{alg:FaNCLacc}), 
which checks the objectives.
Computing $F(\X_t)$ relies only on the observed positions in $\mathbf{\Omega}$ and the singular values of $\X_t$.
Hence, instead of explicitly constructing $\X_t$, 
we maintain 
the SVD 
$\mathbf{U}_t \mathbf{\Sigma}_t \mathbf{V}_t^{\top}$
of $\X_t$ and a sparse matrix $\SO{\X_t}$.
Computing $F(\X_t)$ then takes $O(\NM{\mathbf{\Omega}}{1} r_t)$ time.
Computing $F(\X^a_{t + 1})$ takes $O( \NM{\mathbf{\Omega}}{1} k_t )$ time, as $\mathbf{R}_t$
has rank $k_t$.
Next,
since $\mathbf{Y}_{t}$ is a linear combination of $\X_t$ and $\X_{t - 1}$ in step~4,
we can use the above SVD-factorized form and compute $\NM{\X^a_{t + 1} - \mathbf{Y}_t}{F}^2$ in $O((m + n) k^2_t )$ time.
Thus,
step~7 then takes $O( \NM{\mathbf{\Omega}}{1}k_t + (m + n)k_t^2 )$ time.

Steps~6 and 10 perform inexact proximal step. For the first proximal step
(step~6),
$\mathbf{Y}_t$ (defined in step~4) can be rewritten as $( 1 + \beta_t ) \X_t - \beta_t \X_{t - 1}$,
where $\beta_t = (\alpha_{t - 1} - 1)/\alpha_t$.
When it calls $\textsf{InexactPS}$,
step~1
of Algorithm~\ref{alg:inexactPS} has
\begin{align}
\X_{\text{gd}}
& = \mathbf{Y}_t + \frac{1}{\tau} \SO{\mathbf{Y}_t- \mathbf{O}}
\notag \\
& = (1 + \beta_t) \X_t - \beta_t \X_{t - 1}
+ \frac{1}{\tau} \SO{\mathbf{Y}_t - \mathbf{O}}.
\label{eq:zt}
\end{align}
The first two terms
involve low-rank matrices, while the last term
involves a sparse matrix.
This special ``sparse plus low-rank'' structure 
\cite{mazumder2010spectral}
is essential for the
matrix completion solver,
including the proposed algorithm,
to be efficient.
Specifically,
for any $\mathbf{V} \in \R^{n \times k}$,
$\X_{\text{gd}} \mathbf{V}$ 
can be obtained as
\begin{align}
\X_{\text{gd}} \mathbf{V}
\! = & (1 + \beta_t) \mathbf{U}_t \mathbf{\Sigma}_t (\mathbf{V}_t^{\top} \mathbf{V}) 
- \beta_t \mathbf{U}_{t - 1} \mathbf{\Sigma}_{t - 1} (\mathbf{V}_{t - 1}^{\top} \mathbf{V}) 
\notag \\
& + \frac{1}{\tau}\SO{\mathbf{O} - \mathbf{Y}_t} \mathbf{V}.
\label{eq:temp2}
\end{align}
Similarly,
for any $\mathbf{U} \in \R^{m \times k}$,
$\mathbf{U}^{\top} \X_{\text{gd}}$ 
can be obtained as
\begin{align}
\mathbf{U}^{\top} \X_{\text{gd}}
\! = & (1 + \beta_t) (\mathbf{U}^{\top} \mathbf{U}_t) \mathbf{\Sigma}_t \mathbf{V}_t^{\top} 
\! - \! \beta_t (\mathbf{U}^{\top} \mathbf{U}_{t - 1}) \mathbf{\Sigma}_{t - 1} \mathbf{V}_{t - 1}^{\top} 
\notag \\
& + \frac{1}{\tau} \mathbf{U}^{\top} \SO{\mathbf{O} - \mathbf{Y}_t}.
\label{eq:temp3}
\end{align}
Both \eqref{eq:temp2} and \eqref{eq:temp3} take $O((m + n) k_t k +
\NM{\mathbf{\Omega}}{1} k)$,
instead of $O(m n k)$),
time.
In contrast, existing algorithms for adaptive nonconvex regularizers (such as
\textsf{IRNN} \cite{lu2016nonconvex} and \textsf{GPG} \cite{lu2015generalized}) cannot utilize this special structure
and are slow, as will be demonstrated in Section~\ref{sec:exptmc}.

As $\mathbf{R}_t$ in step~5 of Algorithm~\ref{alg:FaNCLacc}
has $k_t$ columns, each call to 
approximate GSVT takes
$O( (m + n)k_t^2 + \NM{\mathbf{\Omega}}{1} k_t )$ time \cite{quan2015impute}
(instead of $O( m n k_t ) $).
Finally, 
step~5
in Algorithm~\ref{alg:inexactPS}
also takes
$O ( (m + n) k_t^2 + \NM{\mathbf{\Omega}}{1} k_t )$ time.
As a result, 
step~6 of Algorithm~\ref{alg:FaNCLacc}
takes a total of
$O(  (m + n)k_t^2 + \NM{\mathbf{\Omega}}{1} k_t ) $
time.
Step~10 is slightly cheaper 
(as no $\X_{t-1}$ is involved),
and its time complexity is
$O( (m + n)r_t k_t + \NM{\mathbf{\Omega}}{1} r_t )$.
Summarizing, the iteration time complexity of Algorithm~\ref{alg:FaNCLacc} is 
\begin{align}
O( (m + n) k_t^2 + \NM{\mathbf{\Omega}}{1} k_t ).
\label{eq:temp4}
\end{align}
Usually,
$k_t \ll n$ and $\NM{\mathbf{\Omega}}{1} \ll mn$ \cite{candes2009exact,mazumder2010spectral}.
Thus, \eqref{eq:temp4} is much cheaper than
the $O(m n k_t)$ complexity of 
\textsf{FaNCL-acc} on general problems
(Section~\ref{sec:accFaNCL}).

The space complexity is also reduced.
We only need to store the low-rank factorizations of $\X_t$ and $\X_{t-1}$, and the sparse matrices $\SO{\X_t}$ and $\SO{\X_{t-1}}$.
These take a total of $O(  (m + n)k_t + \NM{\mathbf{\Omega}}{1} )$ space (instead of $O(m n)$ in Section~\ref{sec:accFaNCL}).

Note that  these techniques can also be used on 
the simpler non-accelerated \textsf{FaNCL} algorithm (Algorithm~\ref{alg:FaNCL}), 
as discussed in the conference version 
of this paper \cite{yao2015fast}.
It can be
easily shown that its
iteration time complexity is
$O( (m + n) r_t k_t + \NM{\mathbf{\Omega}}{1} r_t )$,
and
its space complexity is $O( (m + n)r_t + \NM{\mathbf{\Omega}}{1})$
(as 
no $\X_{t - 1}$
is involved).

\subsubsection{Comparison with Existing Algorithms}

Table~\ref{tab:timecomp} shows the 
convergence rates,
iteration time complexities, 
and
space complexities of various matrix completion algorithms that
will be empirically compared in Section~\ref{sec:expts}. 
Overall, the proposed 
algorithms (Algorithms~\ref{alg:FaNCL} and \ref{alg:FaNCLacc}) enjoy fast convergence, 
cheap iteration complexity and low memory cost.
While  
Algorithms~\ref{alg:FaNCL} and \ref{alg:FaNCLacc} have
the same
convergence rate,
we will see in Section~\ref{sec:exptmc} that Algorithm~\ref{alg:FaNCLacc} 
(which uses acceleration)
is significantly faster.


\subsection{Robust Principal Component Analysis (RPCA)}
\label{sec:rpca}

Given a noisy data matrix $\mathbf{O} \in \R^{m \times n}$, RPCA assumes that $\mathbf{O}$ can be approximated by the sum of a
low-rank matrix $\X$
plus some sparse noise $\mathbf{S}$ \cite{candes2011robust}. 
Its optimization problem is:
\begin{equation} \label{eq:rpca}
\min_{\X, \mathbf{S}} F(\X, \mathbf{S}) \equiv f(\X, \mathbf{S}) + \lambda r(\X) + \upsilon g(\mathbf{S}),
\end{equation} 
where 
$f(\X, \mathbf{S}) = \frac{1}{2}\NM{\X + \mathbf{S} - \mathbf{O}}{F}^2$,
$r$ is a low-rank regularizer, and $g$ is a sparsity-inducing regularizer.
Here, we allow both $r$ and $g$ to be nonconvex and nonsmooth. Thus, \eqref{eq:rpca} can be seen as a nonconvex extension of RPCA (which uses
the nuclear norm regularizer 
for $r$ 
and 
$\ell_1$-regularizer
for $g$).
Some examples of nonconvex $r$ are shown in Table~\ref{tab:lwregu},
and examples of nonconvex $g$ include the
$\ell_1$-norm,
capped-$\ell_1$-norm \cite{zhang2010analysis} and log-sum-penalty \cite{candes2008enhancing}.

While \eqref{eq:rpca} involves two blocks of parameters ($\X$ and $\mathbf{S}$), they are
not coupled together.  Thus, we can use the separable property 
of proximal operator
\cite{parikh2014proximal}:
\begin{align}
\Prox{\lambda r + \upsilon g}{[ \X, \mathbf{S} ]}
= 
[ 
\Prox{\lambda r}{\X},
\Prox{\upsilon g}{\mathbf{S}}
].
\label{eq:sep}
\end{align}
For many popular sparsity-inducing regularizers, computing $\Prox{\upsilon g}{\mathbf{S}}$
takes only $O(m n)$ time 
\cite{gongZLHY2013}.  For example, when $g(\mathbf{S}) = \sum_{i,j} |S_{ij}|$, 
$[\Prox{\upsilon g}{\mathbf{S}}]_{ij} = \text{sign}(S_{ij}) \max( |S_{ij}| - \upsilon, 0 )$,
where $\text{sign}(x)$ is the sign of $x$. 
However,
directly computing $\Prox{\lambda r}{\X}$ requires $O(m n^2)$ time and is expensive.
To alleviate this problem,
Algorithm~\ref{alg:FaNCLacc}
can be easily extended
to Algorithm~\ref{alg:FaNCrpca}.
The iteration time complexity, which is dominated by the inexact proximal steps  in
steps~6 and 13, is reduced to $O( m n k_t) $.

\begin{algorithm}[ht]
\caption{\textsf{FaNCL-acc} algorithm for RPCA.}
\begin{algorithmic}[1]
\REQUIRE choose $\tau > \rho, \lambda_0 > \lambda, \delta > 0, \nu \in (0,1)$ and $c_1 = \frac{\tau - \rho}{4}$;
	\STATE initialize $\mathbf{V}_0, \mathbf{V}_1 \in \R^{n}$ as random Gaussian matrices,
	$\X_0 = \X_1 = 0$, $\mathbf{S}_0 = \mathbf{S}_1 = 0$ and $\alpha_0 = \alpha_1 = 1$;
	\FOR{$t = 1,2,\dots T$}
	\STATE $\lambda_t = (\lambda_{t-1} -\lambda) \nu + \lambda$;
	\STATE $\begin{bmatrix} \mathbf{Y}_t^{\mathbf{X}} , \mathbf{Y}_t^{\mathbf{S}}\end{bmatrix}
	=
	\begin{bmatrix} \X_t , \mathbf{S}_t \end{bmatrix}$
	
	\quad $+ \frac{\alpha_{t - 1} - 1}{\alpha_t}\left( \begin{bmatrix} \X_t , \mathbf{S}_t \end{bmatrix} 
	- \begin{bmatrix} \X_{t - 1} , \mathbf{S}_{t - 1} \end{bmatrix}\right)$;
	
	\STATE $\mathbf{R}_t = \QR{[\mathbf{V}_t, \mathbf{V}_{t - 1}]}$; 
	\quad\quad // warm start
	\STATE $\X^a_{t + 1} \! = \textsf{InexactPS}(\mathbf{Y}_t^{\mathbf{X}}, \mathbf{R}_t)$;
	\STATE $\mathbf{S}^a_{t + 1} = \Prox{\frac{\upsilon}{\tau} g}{\mathbf{Y}^{\mathbf{S}}_t - \frac{1}{\tau} \nabla_{\mathbf{S}} \f{\mathbf{Y}_t^{\mathbf{X}}, \mathbf{Y}_t^{\mathbf{S}}}}$;
	
	\STATE $\Delta_t = \NM{\X^a_{t + 1} - \mathbf{Y}_t^{\mathbf{S}}}{F}^2 + \NM{\mathbf{S}^a_{t + 1} - \mathbf{Y}_t^{\mathbf{S}}}{F}^2$;
	
	\IF{$F(\X^a_{t + 1}, \mathbf{S}^a_{t + 1}) \le F(\X_t, \mathbf{S}_t) - \frac{\delta}{2} \Delta_t$}
	\STATE $\X_{t + 1} \! = \X^a_{t + 1}$;
	\STATE $\mathbf{S}_{t + 1} = \mathbf{S}^a_{t + 1}$;
	\ELSE
	
	\STATE $\mathbf{X}_{t + 1} \! = \text{\sf InexactPS}(\X_t, \mathbf{R}_t)$;
	\STATE $\mathbf{S}_{t + 1} = \Prox{\frac{\upsilon}{\tau} g}{\mathbf{S}_t - \frac{1}{\tau} \nabla_{\mathbf{S}} \f{\X_t, \mathbf{S}_t}}$;
	\ENDIF
	\STATE $\alpha_{t + 1} = \frac{1}{2} (\sqrt{4\alpha_t^2 + 1} + 1)$;
	\ENDFOR 
	\RETURN $\X_{T + 1}$ and $\mathbf{S}_{T + 1}$.
\end{algorithmic}
\label{alg:FaNCrpca}
\end{algorithm}

Convergence results 
in Section~\ref{sec:accFaNCL}
can be easily extended to this RPCA problem.
Proofs of the following can be found in Appendices~\ref{app:pr:convrpca},
\ref{app:raterpca}, and
\ref{app:thm:convrpca}.

\begin{prop} \label{pr:convrpca}
	The sequence $\left\lbrace [ \X_t, \mathbf{S}_t ] \right\rbrace$ generated from Algorithm~\ref{alg:FaNCrpca} is bounded,
	and has at least one limit point.
\end{prop}

\begin{corollary} \label{cor:raterpca}
	Let $\mathbf{C}_t = \left[ \mathbf{Y}^{\mathbf{X}}_t, \mathbf{Y}^{\mathbf{S}}_t \right]$ if
	steps~10 and 11 are performed,
	and $\mathbf{C}_t = [\X_t, \mathbf{S}_t]$ otherwise.
	Then, $\min_{t = 1, \dots, T} \NM{[\X_{t + 1}, \mathbf{S}_{t + 1}] - \mathbf{C}_t}{F}^2 \! \le \! \frac{F(\X_1, \mathbf{S}_1) - \inf F}{\min(c_1, \delta/2) T}$. 
\end{corollary}

\begin{theorem} \label{thm:convrpca}
Let $\left\lbrace  [\X_{t_j}, \mathbf{S}_{t_j}] \right\rbrace$ be
a subsequence of $\{[\X_t, \mathbf{S}_t]\}$ generated by
Algorithm~\ref{alg:FaNCrpca} such that
$\lim_{t_j \rightarrow \infty} \X_{t_j} = \X_*$
and $\lim_{t_j \rightarrow \infty} \mathbf{S}_{t_j} = \mathbf{S}_*$
With the assumption in Theorem~\ref{the:FaNCL:conv},
$[\X_*, \mathbf{S}_*]$ is a critical point of \eqref{eq:rpca}.
\end{theorem}

The non-accelerated \textsf{FaNCL} (Algorithm~\ref{alg:FaNCL}) can be 
similarly
extended for RPCA.
Same to the accelerated \textsf{FaNCL},
its space complexity is also 
$O( m n k_t )$ 
and 
its per-iteration time complexity
is also
$O( m n k_t )$.


\begin{figure*}[ht]
\centering
\subfigure[$\mathbf{U}^{\top} \mathbf{X}$.]
{\includegraphics[height = 0.21\textwidth]
	{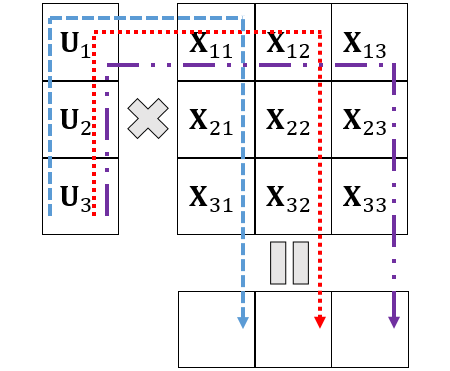}
	\label{fig:parXV}}
\subfigure[$\mathbf{X} \mathbf{V}$.]
{\includegraphics[height = 0.21\textwidth]
	{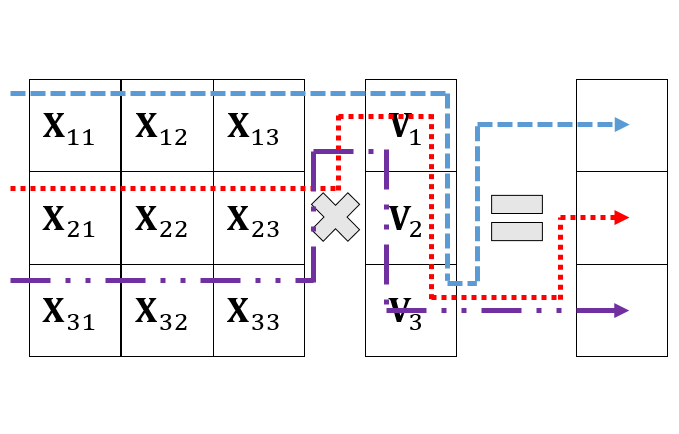}
	\label{fig:parXU}}
\subfigure[Element-wise operation.]
{\includegraphics[height = 0.21\textwidth]
	{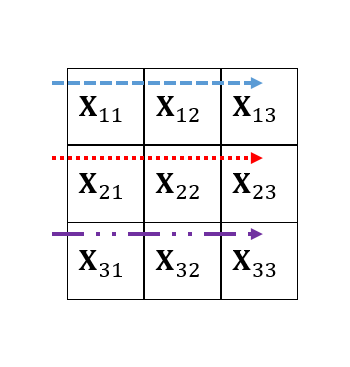}
	\label{fig:elevise}}

\vspace{-10px}
\caption{Parallelization of different matrix operations. Here, the number of
	threads  $q$ is equal to 3. Each dotted path denotes operation of a thread.}
\label{fig:threetype}
\end{figure*}

\section{Parallel \textsf{FaNCL} for Matrix Completion}
\label{sec:parallel}

In this section, we show how the proposed algorithms can be parallelized.
We will only consider the matrix completion problem in \eqref{eq:matcomp}.
Extension to other problems, such as RPCA in Section~\ref{sec:rpca}, can be similarly performed.
Moreover, for simplicity of discussion, we focus on the  simpler
\textsf{FaNCL} algorithm
(Algorithm~\ref{alg:FaNCL}).
Its accelerated variant
(Algorithm~\ref{alg:FaNCLacc}) can be similarly parallelized and is shown in Appendix~C.

Parallel algorithms for matrix completion have been proposed in
\cite{gemulla2011large,yu2012scalable,recht2013parallel}. However,
they are based on stochastic gradient descent and matrix factorization, 
and cannot be directly used here.

\subsection{Proposed Algorithm}
\label{sec:basicidea}

\begin{figure}[ht]
\centering
\subfigure[$\mathbf{U} \mathbf{\Sigma} \mathbf{V}^{\top}$.]
{\includegraphics[height = 0.145\textwidth]{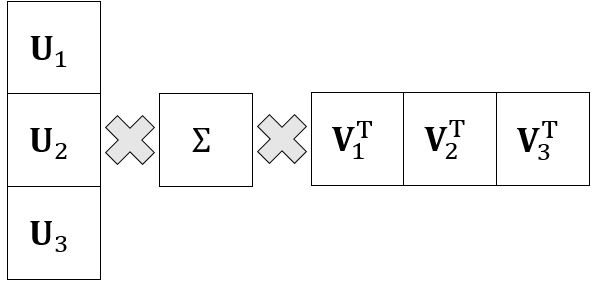} \label{fig:partilr} }
\subfigure[$\mathbf{O}$.]
{\includegraphics[height = 0.145\textwidth]{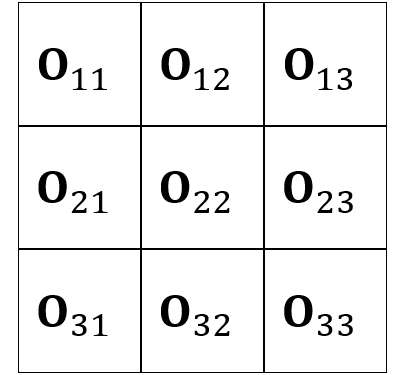} \label{fig:partisp}}

\vspace{-10px}
\caption{Partitioning of variables $\mathbf{U} \mathbf{\Sigma} \mathbf{V}^{\top}$ and $\mathbf{O}$, 
	and three threads are used ($q = 3$).}
\label{fig:parti}
\end{figure}

Operations on a matrix $\mathbf{X}$ are often of the form:
(i) multiplications $\mathbf{U}^{\top} \mathbf{X}$ and $\mathbf{X} \mathbf{V}$ for some
$\mathbf{U}, \mathbf{V}$ (e.g., in \eqref{eq:temp2}, \eqref{eq:temp3});
and
(ii) 
element-wise operation (e.g., evaluation of $F(\mathbf{X})$ in 
\eqref{eq:decrease}).
A popular scheme in parallel linear algebra is block distribution \cite{bertsekas1997parallel}.
Assume that there are $q$ threads for parallelization.
Block distribution partitions
the rows and columns of $\mathbf{X}$ into $q$ parts, leading to a total of
$q^2$ blocks.
For Algorithm~\ref{alg:FaNCL}, 
the most important variables are the low-rank factorized form
$\mathbf{U}_t \mathbf{\Sigma}_t \mathbf{V}_t^{\top}$ of $\X_t$, 
and the sparse matrices $\SO{\X_t}, \SO{\mathbf{O}}$.
Using block distribution, they are simply partitioned as in $q^2$ blocks (Figure~\ref{fig:parti}).
Figure~\ref{fig:threetype} shows how computations of $\mathbf{X} \mathbf{V}$,
$\mathbf{U}^{\top} \mathbf{X}$ and element-wise operation can be easily parallelized. 

The resultant parallelized version of
\textsf{FaNCL} 
is shown in Algorithm~\ref{alg:parallel}.
Steps that can be parallelized 
are marked with ``$\rhd$''.
Two new subroutines are introduced, 
namely,
\textsf{IndeSpan-PL}  
(step~6) which replaces \textsf{QR} factorization, and
\textsf{ApproxGSVT-PL} 
(step~9) which is the parallelized version of Algorithm~\ref{alg:apprGSVT}.
They will be discussed in more detail in the following Sections.
Note that Algorithm~\ref{alg:parallel} is 
equivalent to Algorithm~\ref{alg:FaNCL} except that it is parallelized.
Thus, the
convergence results in Section~\ref{sec:convana} still hold.

\begin{algorithm}[ht]
\caption{\textsf{FaNCL} in parallel: \textsf{FaNCL-PL}.}
\begin{algorithmic}[1]
	\REQUIRE choose $\tau > \rho$, $\lambda_0 > \lambda$ and $\nu \in (0,1)$;
	
	\STATE initialize $\mathbf{V}_0, \mathbf{V}_1 \in \R^{n}$ as random Gaussian matrices and $\X_1 = 0$;
	
	\STATE partition $\X_1$, $\SO{\X_1}$ and $\SO{\mathbf{O}}$;
	
	\STATE start $q$ threads for parallelization;
	
	\FOR{$t = 1,2,\dots T$}
	
	\STATE $\lambda_t = (\lambda_{t-1} -\lambda) \nu^t + \lambda$;
	
	\STATE $\rhd$ $\mathbf{R}_t = \textsf{IndeSpan-PL}\left( [\mathbf{V}_t, \mathbf{V}_{t - 1}] \right)$;
	
	\STATE $\rhd$ $(\X_{\text{gd}})_t = \X_t - \frac{1}{\tau} \SO{\X_t - \mathbf{O}}$;
	
	\FOR{$p = 1, 2,\dots$}
	\STATE $\rhd$ $[ \tilde{\mathbf{X}}_p, \mathbf{R}_t  ]   = 
	\textsf{ApproxGSVT-PL} ( (\X_{\text{gd}})_t, \mathbf{R}_t, \frac{\lambda}{\tau} )$;
	
	\STATE $\rhd$ $a_p = F(  \tilde{\mathbf{X}}_p )$;
	
	\STATE $\rhd$ $a_t \, = F (\X_t)$;
	
	\STATE $\rhd$ $a_F = \| \tilde{\mathbf{X}}_p - \X_t \|_F^2$;
	 
	\STATE \textbf{if} {$a_p \le a_t - c_1 a_F$} break; \textbf{end if}
	
	\ENDFOR
	
	\STATE $\rhd$ $\X_{t + 1} = \tilde{\X}_p$;
	
	\ENDFOR
	
	\RETURN $\X_{T + 1}$.
\end{algorithmic}
\label{alg:parallel}
\end{algorithm}

\subsubsection{Identifying the Span (Step~5)}
\label{sec:qrfact}

In step~4 of Algorithm~\ref{alg:FaNCL},
QR factorization is used 
to find the span of matrix
   $[\mathbf{V}_t, \mathbf{V}_{t - 1}]$.
This can be parallelized with the Householder transformation and Gaussian elimination
\cite{bertsekas1997parallel}, which however
is typically very complex. 
The following Proposition proposes a simpler method to identify the span of a matrix.

\begin{prop} \label{pr:qr}
Given a matrix $\X_{\text{gd}}$,
let the SVD of $\X_{\text{gd}} ^{\top} \X_{\text{gd}}$ be $\mathbf{V} \mathbf{\Sigma}
\mathbf{V}^{\top}$,
$\mathbf{w} = [w_i]$ where $ w_i = \Sigma_{ii}$ if $\Sigma_{ii} > 0$ and $1$ otherwise.
Then, 
$\X_{\text{gd}} \mathbf{V} \left( \Diag{\mathbf{w}} \right)^{-\frac{1}{2}}$ is orthogonal and contains $\Span{\X_{\text{gd}}}$.
\end{prop}

The resultant parallel algorithm is shown in Algorithm~\ref{alg:iden}.
Its time complexity is $O( (\frac{n}{q} + q)k^2 + k^3 )$.
Algorithm~\ref{alg:parallel} calls Algorithm~\ref{alg:iden} 
with 
input
$[\mathbf{V}_t, \mathbf{V}_{t - 1}]$, and
thus takes
$O( (\frac{n}{q} + q)k_t^2 + k_t^3 )$
time,
where $k_t = r_t + r_{t - 1}$.   
We do not parallelize steps~2-4,
	as only 
	$k \times k$ 
	matrices 
	are involved and $k$ is small. Moreover,
though step~3 uses SVD, it only takes $O(k^3)$ time.

\begin{algorithm}[ht]
\caption{Parallel algorithm to identify the span of $\X_{\text{gd}}$: $\textsf{IndeSpan-PL}(\X_{\text{gd}})$.}
\begin{algorithmic}[1]
	\REQUIRE matrix $\X_{\text{gd}} \in \R^{n \times k}$;
	\STATE $\rhd$ $\mathbf{B} = \X_{\text{gd}}^{\top} \X_{\text{gd}}$;
	\STATE $[\mathbf{U}, \mathbf{\Sigma}, \mathbf{V}] = \textsf{SVD}(\mathbf{B})$; 
	\STATE construct $\mathbf{w}$ as in Proposition~\ref{pr:qr};
	\STATE $\mathbf{V} = \mathbf{V} \Diag{\mathbf{w}}$;
	\STATE $\rhd$ $\mathbf{Q} = \X_{\text{gd}} \mathbf{V} $;
	\RETURN $\mathbf{Q}$. // $\mathbf{Q} = [\mathbf{Q}_1^{\top}, \dots, \mathbf{Q}_p^{\top}]^{\top}$
\end{algorithmic}
\label{alg:iden}
\end{algorithm} 

\subsubsection{Approximate GSVT (Step~8)}
\label{sec:inexactps}

The key steps in approximate GSVT (Algorithm~\ref{alg:apprGSVT}) are the power method and SVD.
The power method can be parallelized straightforwardly as in
Algorithm~\ref{alg:powermethodPL}, in which we also replace the \textsf{QR} subroutine with Algorithm~\ref{alg:iden}.

\begin{algorithm}[ht]
\caption{Parallel power method:
		\textsf{Powermethod-PL}$(\mathbf{Z}, \mathbf{R})$.}
	\begin{algorithmic}[1]
		\REQUIRE matrix $\mathbf{Z} \in \R^{m \times n}$, $\mathbf{R} \in \R^{n \times k}$.
		\STATE $\rhd$ $\mathbf{Y}_1 = \mathbf{Z} \mathbf{R}$;
		\FOR{$j = 1, 2, \dots, J$}
		\STATE $\rhd$ $\mathbf{Q}_{j} = \textsf{IndeSpan-PL}(\mathbf{Y}_j)$;  
		\STATE $\rhd$ $\mathbf{Y}_{j + 1} = \mathbf{Z} (\mathbf{Z}^{\top} \mathbf{Q}_{j})$;
		\ENDFOR
		\RETURN $\mathbf{Q}_{J}$. 
	\end{algorithmic}
	\label{alg:powermethodPL}
\end{algorithm}

As for SVD, multiple QR factorizations are usually needed for parallelization \cite{bertsekas1997parallel}, 
	which are complex as discussed in Section~\ref{sec:qrfact}.
The following Proposition performs it in a simpler manner.

\begin{prop} \label{pr:rsvd}
Given 
a matrix $\mathbf{B} \in \R^{n \times k}$,
let $\mathbf{P}  \in \R^{n \times k}$ be orthogonal and equals $\Span{\mathbf{B}}$, 
and the SVD of $\mathbf{P}^{\top} \mathbf{B}$ be $\mathbf{U} \mathbf{\Sigma} \mathbf{V}^{\top}$.
Then, 
the SVD of $\mathbf{B}$ is
$(\mathbf{P} \mathbf{U})\mathbf{\Sigma}\mathbf{V}$.
\end{prop}

The resultant parallelized procedure for approximate GSVT is shown in
Algorithm~\ref{alg:apprGSVTPl}.
At step~5,
a small  SVD is performed (by a single thread)
on  the $k \times k$ matrix
	$\mathbf{P}^{\top} \mathbf{B}$.
At step~8 of Algorithm~\ref{alg:parallel},
$\tilde{\X}_p$ is returned from Algorithm~\ref{alg:apprGSVTPl},
and we keep $\tilde{\X}_p$ in its low-rank factorized form.
Besides, when Algorithm~\ref{alg:apprGSVTPl} is called,
$\mathbf{Z} = (\X_{\text{gd}})_t$
and has the ``sparse plus low-rank'' structure mentioned earlier. Hence, \eqref{eq:temp2} and \eqref{eq:temp3} can be used to speed up
matrix multiplications\footnote{As no acceleration is used, $\beta_t$ 
in \eqref{eq:temp2} and \eqref{eq:temp3} 
is equal to zero
in these two equations.}.
As $\mathbf{R}_t$ has $k_t$ columns in Algorithm~\ref{alg:parallel},
\textsf{PowerMethod-PL} in step~1 takes
$O( \frac{k_t}{q} \NM{\mathbf{\Omega}}{1} + \frac{m + n}{q} k_t^2 )$ time, 
steps~2-6 take $O( (\frac{n}{q} + q)k_t^2 + k_t^3 )$ time,
and the rest takes $O(k_t)$ time.
The total time complexity 
for Algorithm~\ref{alg:apprGSVTPl} 
is 
$O( \frac{k_t}{q} \NM{\mathbf{\Omega}}{1} + \frac{m + n}{q} k_t^2 
+ (q + k_t)k_t^2 )$.

\begin{algorithm}[ht]
\caption{Approximate GSVT in parallel:
	\textsf{ApproxGSVT-PL}$(\mathbf{Z}, \mathbf{R}, \mu)$.}
\begin{algorithmic}[1]
	\REQUIRE partitioned matrix $\mathbf{Z} \in \R^{m \times n}$ and $\mathbf{R} \in \R^{n \times k}$;
	
	\STATE $\rhd$ $\mathbf{Q} = \text{\sf PowerMethod-PL}(\mathbf{Z}, \mathbf{R})$;
	
	\STATE $\rhd$ $\mathbf{B} = \mathbf{Z}^{\top} \mathbf{Q}$; 
	// $\mathbf{B} \in \R^{n \times k}$
	
	\STATE $\rhd$ $\mathbf{P} = \textsf{Iden-Span}(\mathbf{B})$;
	
	\STATE $\rhd$ $\X_{\text{gd}} = \mathbf{P}^{\top} \mathbf{B}$;
	
	\STATE $[\mathbf{U}, \mathbf{\Sigma}, \mathbf{V}] = \textsf{SVD}(\X_{\text{gd}})$;
	// $\mathbf{U}, \mathbf{\Sigma}, \mathbf{V}, \X_{\text{gd}} \in \R^{k \times k}$
	
	\STATE $\rhd$ $\mathbf{U} = \mathbf{P} \mathbf{U}$;
	
	\STATE $a = $ number of $\Sigma_{ii}$'s that are $>\gamma$ in Corollary~\ref{cor:threh};
	
	\STATE $\rhd$ $\mathbf{U}_a = a$ leading columns of $\mathbf{U}$;
	
	\STATE $\rhd$ $\mathbf{V}_a = a$ leading columns of $\mathbf{V}$;
	
	\FOR{$i = 1,2,\dots,a$}
	\STATE obtain $y^*_i$ from \eqref{eq:proRed};
	\ENDFOR
	
	\RETURN  
	the low-rank components of $\tilde{\mathbf{X}}$
	($\mathbf{Q} \mathbf{U}_a$, $\Diag{[ y_1^*,\dots,y_{a}^* ]}$ and $\mathbf{V}_a^{\top}$),
	and $\mathbf{V}$.
\end{algorithmic}
\label{alg:apprGSVTPl}
\end{algorithm}

\subsubsection{Checking of Objectives (steps~9-11)}
\label{sec:chkobj}

As shown in Figures~\ref{fig:elevise},
computation of $\NM{\SO{\mathbf{X}_t - \mathbf{O}}}{F}^2$ in $F(\cdot)$
can be directly parallelized and takes 
$O(\frac{1}{p}\NM{\mathbf{\Omega}}{1})$ time.
As $r$ only relies on $\mathbf{\Sigma}_t$, 
only one thread is needed to evaluate $r(\X_t)$.
Thus, computing $F(\X_t)$ takes $O( \frac{r_t}{q} \NM{\mathbf{\Omega}}{1} )$ time.
Similarly,
computing $F( \tilde{\mathbf{X}}_p )$
takes $O( \frac{k_t}{q} \NM{\mathbf{\Omega}}{1} )$ time.
As $\NM{\tilde{\X}_p - \mathbf{X}_t}{F}^2 
= \Tr{\tilde{\X}_p^{\top} \tilde{\X}_p - 2 \tilde{\mathbf{X}}_p^{\top} \mathbf{X}_t -
\mathbf{X}_t^{\top} \mathbf{X}_t}$, the low-rank factorized forms of $\tilde{\mathbf{X}}_p$ and $\mathbf{X}_t$ can be utilized.
From Figures~\ref{fig:parXV} and 
\ref{fig:parXU}, 
it can be performed in $O(\frac{m + n}{q} k_t^2 )$ time.
The time complexity for steps~9-11 in Algorithm~\ref{alg:parallel} is
$O(\frac{k_t}{q}\NM{\mathbf{\Omega}}{1}
+ \frac{m + n}{q}k_t^2)$.
The iteration time complexity for Algorithm~\ref{alg:parallel} is thus
$O( 
\frac{1}{q}((m + n) k_t^2
+ 
\NM{\mathbf{\Omega}}{1} k_t)
+ 
(q + k_t) k_t^2)$.
Compared with \eqref{eq:temp4},
the speedup w.r.t. the number of threads $q$
is almost linear.

\begin{table*}[ht]
	\centering
	\caption{$\!\!$ Matrix completion performance on the synthetic data.
		NMSE is scaled by $10^{-2}$, CPU time is in seconds and the number in brackets is data sparsity.}
	\vspace{-10px}
	\begin{tabular}{cc|ccc|ccc|ccc}
		\hline
		&                     &       \multicolumn{3}{c|}{$m=500$ ($12.43\%$)}       &       \multicolumn{3}{c|}{$m=1000$ ($6.91\%$)}       &       \multicolumn{3}{c}{$m=2000$ ($3.80\%$)}        \\
		&                     &          NMSE          & rank &         time         &          NMSE          & rank &         time         &          NMSE          & rank &         time         \\ \hline
		nuclear norm   &    \textsf{APG}     &     4.26$\pm$0.01      &  50  &     12.6$\pm$0.7     &     4.27$\pm$0.01      &  61  &     99.6$\pm$9.1     &     4.13$\pm$0.01      &  77  &   1177.5$\pm$134.2   \\ \cline{2-11}
		& \textsf{AIS-Impute} &     4.11$\pm$0.01      &  55  &     5.8$\pm$2.9      &     4.01$\pm$0.03      &  57  &     37.9$\pm$2.9     &     3.50$\pm$0.01      &  65  &    338.1$\pm$54.1    \\ \cline{2-11}
		&   \textsf{active}   &     5.37$\pm$0.03      &  53  &     12.5$\pm$1.0     &     6.63$\pm$0.03      &  69  &     66.4$\pm$3.3     &     6.44$\pm$0.10      &  85  &    547.3$\pm$91.6    \\ \hline\hline
		fixed rank    &   \textsf{LMaFit}   &     3.08$\pm$0.02      &  5   &     0.5$\pm$0.1      &     3.02$\pm$0.02      &  5   &     1.3$\pm$0.1      &     2.84$\pm$0.03      &  5   &     4.9$\pm$0.3      \\ \cline{2-11}
		&   \textsf{ER1MP}    &     21.75$\pm$0.05     &  40  &     0.3$\pm$0.1      &     21.94$\pm$0.09     &  54  &     0.8$\pm$0.1      &     20.38$\pm$0.06     &  70  &     2.5$\pm$0.3      \\ \hline\hline
		capped $\ell_1$ &    \textsf{IRNN}    & \textbf{1.98$\pm$0.01} &  5   &     14.5$\pm$0.7     &     1.99$\pm$0.01      &  5   &    146.0$\pm$2.6     & \textbf{1.79$\pm$0.01} &  5   &   2759.9$\pm$252.8   \\ \cline{2-11}
		&    \textsf{GPG}     & \textbf{1.98$\pm$0.01} &  5   &     14.8$\pm$0.9     &     1.99$\pm$0.01      &  5   &    144.6$\pm$3.1     & \textbf{1.79$\pm$0.01} &  5   &   2644.9$\pm$358.0   \\ \cline{2-11}
		&   \textsf{FaNCL}    & \textbf{1.97$\pm$0.01} &  5   &     0.3$\pm$0.1      &     1.98$\pm$0.01      &  5   &     1.0$\pm$0.1      & \textbf{1.79$\pm$0.01} &  5   &     5.0$\pm$0.4      \\ \cline{2-11}
		& \textsf{FaNCL-acc}  & \textbf{1.97$\pm$0.01} &  5   & \textbf{0.1$\pm$0.1} & \textbf{1.95$\pm$0.01} &  5   & \textbf{0.5$\pm$0.1} & \textbf{1.78$\pm$0.01} &      & \textbf{2.3$\pm$0.2} \\ \hline
		LSP       &    \textsf{IRNN}    & \textbf{1.96$\pm$0.01} &  5   &     16.8$\pm$0.6     & \textbf{1.89$\pm$0.01} &  5   &    196.1$\pm$3.9     & \textbf{1.79$\pm$0.01} &  5   &   2951.7$\pm$361.3   \\ \cline{2-11}
		&    \textsf{GPG}     & \textbf{1.96$\pm$0.01} &  5   &     16.5$\pm$0.4     & \textbf{1.89$\pm$0.01} &  5   &    193.4$\pm$2.1     & \textbf{1.79$\pm$0.01} &  5   &   2908.9$\pm$358.0   \\ \cline{2-11}
		&   \textsf{FaNCL}    & \textbf{1.96$\pm$0.01} &  5   &     0.4$\pm$0.1      & \textbf{1.89$\pm$0.01} &  5   &     1.3$\pm$0.1      & \textbf{1.79$\pm$0.01} &  5   &     5.5$\pm$0.4      \\ \cline{2-11}
		& \textsf{FaNCL-acc}  & \textbf{1.96$\pm$0.01} &  5   & \textbf{0.2$\pm$0.1} & \textbf{1.89$\pm$0.01} &  5   & \textbf{0.7$\pm$0.1} & \textbf{1.77$\pm$0.01} &      & \textbf{2.4$\pm$0.2} \\ \hline
		TNN       &    \textsf{IRNN}    & \textbf{1.96$\pm$0.01} &  5   &     18.8$\pm$0.6     & \textbf{1.88$\pm$0.01} &  5   &    223.1$\pm$4.9     & \textbf{1.77$\pm$0.01} &  5   &   3220.3$\pm$379.7   \\ \cline{2-11}
		&    \textsf{GPG}     & \textbf{1.96$\pm$0.01} &  5   &     18.0$\pm$0.6     & \textbf{1.88$\pm$0.01} &  5   &    220.9$\pm$4.5     & \textbf{1.77$\pm$0.01} &  5   &   3197.8$\pm$368.9   \\ \cline{2-11}
		&   \textsf{FaNCL}    & \textbf{1.95$\pm$0.01} &  5   &     0.4$\pm$0.1      & \textbf{1.88$\pm$0.01} &  5   &     1.4$\pm$0.1      & \textbf{1.77$\pm$0.01} &  5   &     6.1$\pm$0.5      \\ \cline{2-11}
		& \textsf{FaNCL-acc}  & \textbf{1.96$\pm$0.01} &  5   & \textbf{0.2$\pm$0.1} & \textbf{1.88$\pm$0.01} &  5   & \textbf{0.8$\pm$0.1} & \textbf{1.77$\pm$0.01} &      & \textbf{2.9$\pm$0.2} \\ \hline
	\end{tabular}
	\label{tab:sythmatcomp}
\end{table*}


\section{Experiments}
\label{sec:expts}

In this section, we perform experiments on matrix completion, RPCA  
and the parallelized variant of Algorithm~\ref{alg:FaNCLacc}.
We use a Windows server 2013 system with Intel Xeon E5-2695-v2 CPU (12 cores, 2.4GHz) and 256GB memory. 
All the
algorithms in Sections~\ref{sec:exptmc} and \ref{sec:expt2} are implemented in Matlab.
For Section~\ref{sec:expparallel},
we use C++, the Intel-MKL package
for matrix operations, and the standard thread library
for multi-thread programming.


\subsection{Matrix Completion}
\label{sec:exptmc}

We compare a number of low-rank matrix completion solvers, including
models based on 
(i) the commonly used (convex) nuclear norm regularizer; 
(ii) fixed-rank factorization models \cite{wen2012solving,wang2015orthogonal},
which decompose the  observed matrix $\mathbf{O}$ into a product of 
rank-$k$ matrices $\mathbf{U}$ and $\mathbf{V}$. Its optimization problem can be written as:
$\min_{U,V}\frac{1}{2}\NM{\SO{\mathbf{U} \mathbf{V} - \mathbf{O}}}{F}^2+ \frac{\lambda}{2}(\NM{\mathbf{U}}{F}^2 + \NM{\mathbf{V}}{F}^2)$;
and 
(iii) nonconvex regularizers, including the capped-$\ell_1$ (with $\theta$ in
Table~\ref{tab:lwregu} set to $2 \lambda$), LSP
(with $\theta = \sqrt{\lambda}$), and TNN
(with $\theta = 3$).

The nuclear norm minimization algorithms to be compared include:
(i) Accelerated proximal gradient (\textsf{APG}) algorithm \cite{ji2009accelerated,toh2010accelerated}, 
with the partial SVD computed by the PROPACK package \cite{larsen1998lanczos};
(ii) \textsf{AIS-Impute} \cite{mazumder2010spectral}, an inexact and accelerated proximal algorithm.
The ``sparse plus low-rank'' structure of  the
matrix iterate is utilized to speed up computation (Section~\ref{sec:matcomp}); and
(iii) Active subspace selection 
(denoted ``\textsf{active}'') \cite{hsieh2014nuclear}, which adds/removes rank-one subspaces from the active set in each iteration.  
as they have been shown to be less efficient 
\cite{hsieh2014nuclear,quan2015impute}.
For fixed-rank factorization models
(where the rank is tuned by the validation set),
we compare with the two state-of-the-art algorithms:
(i) Low-rank matrix fitting (\textsf{LMaFit})
	algorithm
	\cite{wen2012solving}; and
(ii) Economical rank-one matrix pursuit (\textsf{ER1MP}) \cite{wang2015orthogonal},
	which pursues a rank-one basis in each iteration.
We do not compare with the concave-convex procedure \cite{zhang2010analysis,hu2013fast}, since
it has been shown to be inferior to \textsf{IRNN} \cite{gongZLHY2013}.
For models with nonconvex low-rank regularizers, 
we compare with the following solvers:
(i) Iterative reweighted nuclear norm (\textsf{IRNN})	\cite{lu2016nonconvex};
(ii) Generalized proximal gradient (\textsf{GPG}) algorithm \cite{lu2015generalized}, with the
	underlying problem (\ref{eq:proRed}) solved using the closed-form solutions
	in \cite{gongZLHY2013}; and
(iii) The proposed \textsf{FaNCL} algorithm (Algorithm~\ref{alg:FaNCL}) and its
	accelerated variant \textsf{\textsf{FaNCL}-acc} 
	(Algorithm~\ref{alg:FaNCLacc}).
	We set $J = 3$ and $p = 1$.

All the algorithms are  stopped when the relative difference
in objective values between consecutive iterations is smaller than 
${10}^{-4}$.

\subsubsection{Synthetic Data}
\label{sec:matcomp:syn}

The observed $m\times m$ matrix is generated as 
$\mathbf{O} = \mathbf{U} \mathbf{V} + \mathbf{G}$,  where 
the elements 
of $\mathbf{U} \in \R^{m \times k}, \mathbf{V} \in \R^{k \times m}$ 
(with $k = 5$) are 
sampled i.i.d. from the standard normal distribution $\mathcal{N}(0,
1)$, and
elements of $\mathbf{G}$ sampled from $\mathcal{N}(0, 0.1)$.
A total of $\NM{\mathbf{\Omega}}{1} = 2 m k \log(m)$ random elements
in $\mathbf{O}$ are observed.  
Half of them are used for training, and the rest as validation
set for parameter tuning.
Testing is performed on the unobserved elements.

For performance evaluation,  
we use (i) the normalized mean squared error:
$\text{NMSE} = \NM{P_{\mathbf{\Omega}^{\bot}}(\mathbf{X} - \mathbf{U V})}{F} 
/ \NM{P_{\mathbf{\Omega}^{\bot}}(\mathbf{U V})}{F}$,
where $\mathbf{X}$ is the recovered matrix and
$\mathbf{\Omega}^{\bot}$
denotes the unobserved positions;
(ii) rank of $\mathbf{X}$;
and 
(iii) training CPU time.
We vary $m$ in $\{500, 1000, 2000\}$.
Each experiment is repeated five times.

Results
are shown in 
\footnote{For all tables in the sequel,
	the best and comparable results (according to the pairwise t-test with 95\% confidence) are highlighted.}
Table~\ref{tab:sythmatcomp}.
As can be seen, nonconvex regularization (capped-$\ell_1$, LSP and TNN) leads to much lower NMSE's than 
convex nuclear norm regularization and fixed-rank factorization.
Moreover, 
nuclear norm regularization
and \textsf{ER1MP} produce much higher ranks.
In terms of speed
among the nonconvex low-rank solvers,
\textsf{FaNCL} is faster than 
\textsf{GPG} and \textsf{IRNN},
while \textsf{FaNCL-acc} is even faster. Moreover,
the larger the matrix, the higher are the speedups of \textsf{FaNCL} and \textsf{FaNCL-acc}
over \textsf{GPG} and \textsf{IRNN}.

\begin{figure}[ht]
\centering
\subfigure[capped-$\ell_1$.]{\includegraphics[width =
	0.24\textwidth]{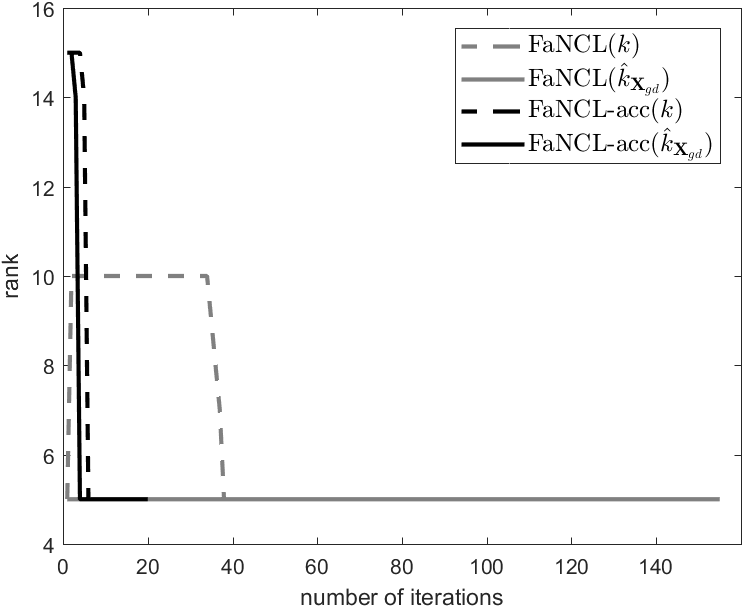}}
\subfigure[LSP.]{\includegraphics[width =
	0.24\textwidth]{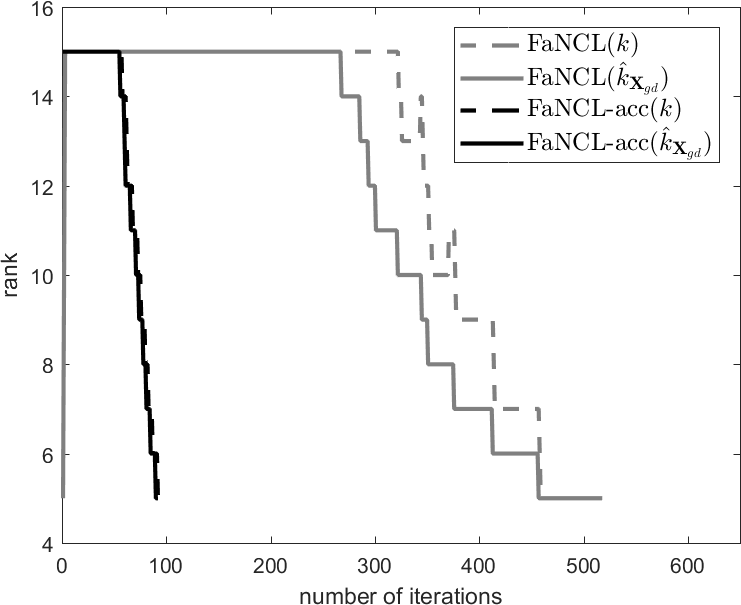}}

\vspace{-10px}
\caption{$k$ and $\hat{k}_{\X_{\text{gd}}}$ 
	vs the number of iterations 
	on the synthetic data set with $m = 500$.
	The plot of TNN is similar and thus not shown.}
\label{fig:rankmatcomp}
\end{figure}

Next,
we demonstrate 
the ability 
of \textsf{FaNCL} and \textsf{FaNCL-acc}
in maintaining low-rank iterates.
Figure~\ref{fig:rankmatcomp} shows 
$k$
(the rank
of $\mathbf{R}_t$)
and 
$\hat{k}_{\mathbf{X}_+}$
(the rank of $\X_{t + 1}$)
vs the number of iterations for $m=500$.
As can be seen, $k \ge \hat{k}_{\X_{\text{gd}}}$, which agrees with the assumption in 
Proposition~\ref{pr:apprGSVT}.
Besides,
as \textsf{FaNCL/FaNCL-acc} converges,
$k$ and $\hat{k}_{\X_{\text{gd}}}$ gradually converge to the same value. Moreover,
recall that the data matrix is of size $500 \times 500$.
Hence, the ranks of the iterates obtained by both algorithms are low.

\subsubsection{Recommendation Data sets}
\label{sec:recsys}

\noindent
\textbf{MovieLens}:
First, we perform
experiments on the popular
\textit{MovieLens}
data set
(Table~\ref{tab:recSys}),
which contain ratings of different users on movies.
We follow the setup in \cite{wang2015orthogonal},  
and use $50\%$ of the observed ratings for training, $25\%$ for validation and the rest for testing.
For performance evaluation, we use the root mean squared error on the test set
$\bar{\mathbf{\Omega}}$:
$\text{RMSE} = \sqrt{ \NM{\mathcal{P}_{\bar{\mathbf{\Omega}}}(\mathbf{X} - \mathbf{O})}{F}^2 / \NM{\mathbf{\bar{\Omega}}}{1}}$,
where $\mathbf{X}$ is the recovered matrix.
The experiment is repeated five times.

\begin{table}[ht]
\centering
\caption{Recommendation data sets used in the experiments.}
\vspace{-10px}
\begin{tabular}{cc | c | c | c }
\hline
                &                  & \#users & \#movies & \#ratings   \\ \hline
\textit{MovieLens} &  \textit{100K}   & 943     & 1,682    & 100,000     \\ \cline{2-5}
                &   \textit{1M}    & 6,040   & 3,449    & 999,714     \\ \cline{2-5}
                &   \textit{10M}   & 69,878  & 10,677   & 10,000,054  \\ \hline
\multicolumn{2}{c|}{\textit{netflix}} & 480,189 & 17,770   & 100,480,507 \\ \hline
\multicolumn{2}{c|}{\textit{yahoo}}  & 249,012 & 296,111  & 62,551,438  \\ \hline
\end{tabular}
\label{tab:recSys}
\end{table}

\begin{figure}[ht]
\centering
\subfigure[capped-$\ell_1$. \label{fig:cap}]
{\includegraphics[width = 0.24\textwidth]{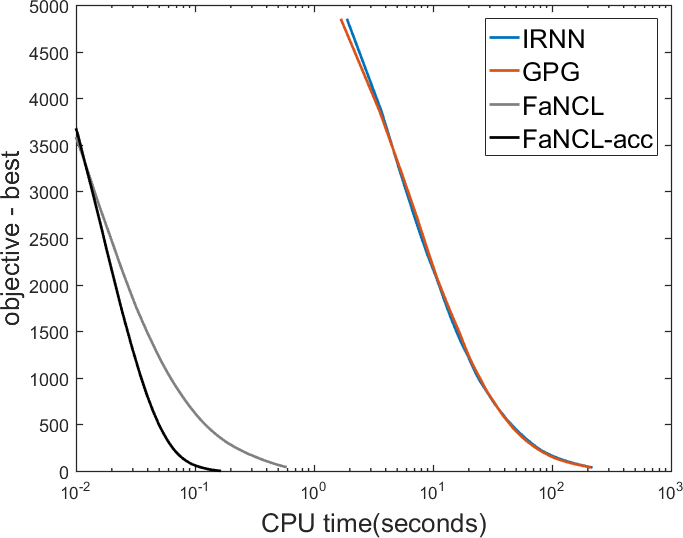}}
\subfigure[LSP. \label{fig:lsp}]
{\includegraphics[width = 0.24\textwidth]{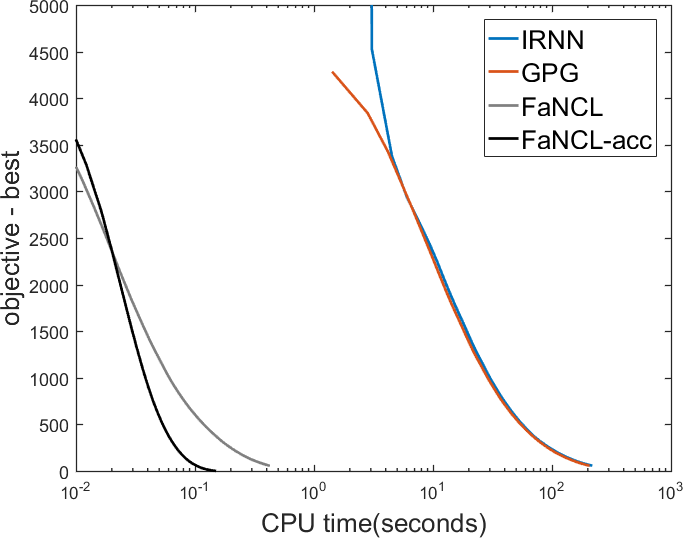}}

\vspace{-10px}
\caption{Objective vs CPU time for the capped-$\ell_1$ and LSP on 
\textit{MovieLens-100K}.  The plot of TNN is similar and thus not shown.}
\label{fig:movielen:obj}
\end{figure}

\begin{figure}[ht]
\centering

\subfigure[\textit{1M}.]
{\includegraphics[width = 0.24 \textwidth]{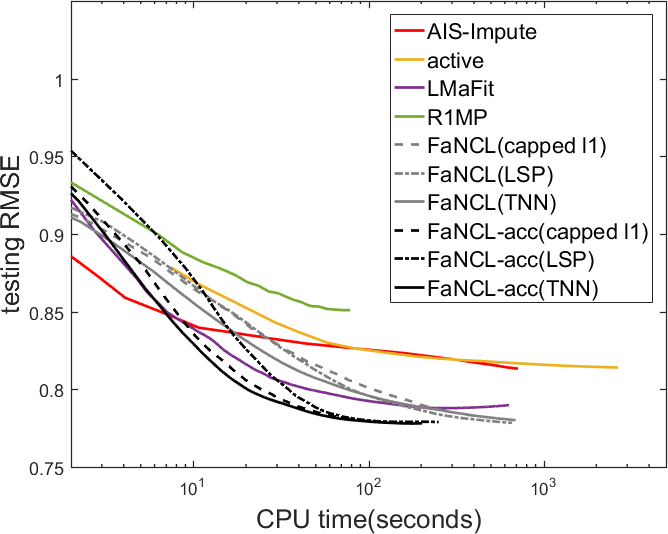}}
\subfigure[\textit{10M}.]
{\includegraphics[width = 0.24 \textwidth]{figures/recsys/10M-all}}

\vspace{-10px}
\caption{RMSE vs CPU time on the \textit{MovieLens-1M} and \textit{10M} data sets.}
\label{fig:movie:rmse}
\end{figure}

Results are shown in Table~\ref{tab:movielen}. 
Again, nonconvex regularizers lead to the lowest RMSE's. 
Moreover, \textsf{FaNCL-acc}
is also the fastest among nonconvex low-rank solvers,
even faster than the state-of-the-art 
\textsf{GPG}.
In particular, 
\textsf{FaNCL} and its accelerated variant \textsf{FaNCL-acc}
are the only solvers (for nonconvex regularization) 
that can be run on the
\textit{MovieLens-1M} and \textit{10M} data sets.
Figure~\ref{fig:movielen:obj} compares the objectives vs CPU time 
for the nonconvex regularization solvers
on \textit{MovieLens-100K}.
As can be seen,
\textsf{FaNCL} and \textsf{FaNCL-acc} decrease the objective and RMSE much faster than the others.
Figure~\ref{fig:movie:rmse} 
shows the testing RMSEs on \textit{MovieLens-1M} and \textit{10M}.
As can be seen, \textsf{FaNCL-acc} is the fastest.

\begin{table*}[ht]
\centering
\caption{Matrix completion results on the \textit{MovieLens} data sets.
	Methods are too slow to run are indicated as ``---''.
}
\vspace{-10px}

\label{tab:movielen}
\begin{tabular}{cc|cc|cc|cc|cc|cc}
	\hline
	         &                     & \multicolumn{2}{c|}{\textit{MovieLens-100K}} & \multicolumn{2}{c|}{\textit{MovieLens-1M}} & \multicolumn{2}{c|}{\textit{MovieLens-10M}} & \multicolumn{2}{c}{\textit{netflix}} & \multicolumn{2}{c}{\textit{yahoo}} \\
	         &                     &           RMSE           &       rank        &           RMSE           &      rank       &           RMSE           &       rank       &           RMSE           &   rank    &           RMSE           &  rank   \\ \hline
	nuclear  &    \textsf{APG}     &     0.877$\pm$0.001      &        36         &     0.818$\pm$0.001      &       67        &           ---            &       ---        &           ---            &    ---    &           ---            &   ---   \\ \cline{2-12}
	  norm   & \textsf{AIS-Impute} &     0.878$\pm$0.002      &        36         &     0.819$\pm$0.001      &       67        &     0.813$\pm$0.001      &       100        &           ---            &    ---    &           ---            &   ---   \\ \cline{2-12}
	         &   \textsf{active}   &     0.878$\pm$0.001      &        36         &     0.820$\pm$0.001      &       67        &     0.814$\pm$0.001      &       100        &           ---            &    ---    &           ---            &   ---   \\ \hline\hline
	 fixed   &   \textsf{LMaFit}   &     0.865$\pm$0.002      &         2         &     0.806$\pm$0.003      &        6        &     0.792$\pm$0.001      &        9         &     0.811$\pm$0.001      &    15     &     0.666$\pm$0.001      &   10    \\ \cline{2-12}
	  rank   &   \textsf{ER1MP}    &     0.917$\pm$0.003      &         5         &     0.853$\pm$0.001      &       13        &     0.852$\pm$0.002      &        22        &     0.862$\pm$0.006      &    25     &     0.810$\pm$0.003      &   77    \\ \hline\hline
	 capped  &    \textsf{IRNN}    & \textbf{0.854$\pm$0.003} &         3         &           ---            &       ---       &           ---            &       ---        &           ---            &    ---    &           ---            &   ---   \\ \cline{2-12}
	$\ell_1$ &    \textsf{GPG}     & \textbf{0.855$\pm$0.002} &         3         &           ---            &       ---       &           ---            &       ---        &           ---            &    ---    &           ---            &   ---   \\ \cline{2-12}
	         &   \textsf{FaNCL}    & \textbf{0.855$\pm$0.003} &         3         & \textbf{0.788$\pm$0.002} &        5        &     0.783$\pm$0.001      &        8         &     0.798$\pm$0.001      &    13     &     0.656$\pm$0.001      &    8    \\ \cline{2-12}
	         & \textsf{FaNCL-acc}  &     0.860$\pm$0.009      &         3         &     0.791$\pm$0.001      &        5        & \textbf{0.778$\pm$0.001} &        8         &     0.795$\pm$0.001      &    13     &     0.651$\pm$0.001      &    8    \\ \hline
	  LSP    &    \textsf{IRNN}    &     0.856$\pm$0.001      &         2         &           ---            &       ---       &           ---            &       ---        &           ---            &    ---    &           ---            &   ---   \\ \cline{2-12}
	         &    \textsf{GPG}     &     0.856$\pm$0.001      &         2         &           ---            &       ---       &           ---            &       ---        &           ---            &    ---    &           ---            &   ---   \\ \cline{2-12}
	         &   \textsf{FaNCL}    &     0.856$\pm$0.001      &         2         & \textbf{0.786$\pm$0.001} &        5        & \textbf{0.779$\pm$0.001} &        9         &     0.794$\pm$0.001      &    15     &     0.652$\pm$0.001      &    9    \\ \cline{2-12}
	         & \textsf{FaNCL-acc}  & \textbf{0.853$\pm$0.001} &         2         &     0.787$\pm$0.001      &        5        & \textbf{0.779$\pm$0.001} &        9         & \textbf{0.792$\pm$0.001} &    15     & \textbf{0.650$\pm$0.001} &    8    \\ \hline
	  TNN    &    \textsf{IRNN}    & \textbf{0.854$\pm$0.004} &         3         &           ---            &       ---       &           ---            &       ---        &           ---            &    ---    &           ---            &   ---   \\ \cline{2-12}
	         &    \textsf{GPG}     & \textbf{0.853$\pm$0.005} &         3         &           ---            &       ---       &           ---            &       ---        &           ---            &    ---    &           ---            &   ---   \\ \cline{2-12}
	         &   \textsf{FaNCL}    &     0.865$\pm$0.016      &         3         & \textbf{0.786$\pm$0.001} &        5        &     0.780$\pm$0.001      &        8         &     0.797$\pm$0.001      &    13     &     0.657$\pm$0.001      &    7    \\ \cline{2-12}
	         & \textsf{FaNCL-acc}  &     0.861$\pm$0.009      &         3         & \textbf{0.786$\pm$0.001} &        5        & \textbf{0.778$\pm$0.001} &        9         &     0.795$\pm$0.001      &    13     & \textbf{0.650$\pm$0.001} &    7    \\ \hline
\end{tabular}
\end{table*}

\begin{figure}[ht]
	\centering
	\subfigure[capped-$\ell_1$.]
	{\includegraphics[width = 0.24\textwidth]{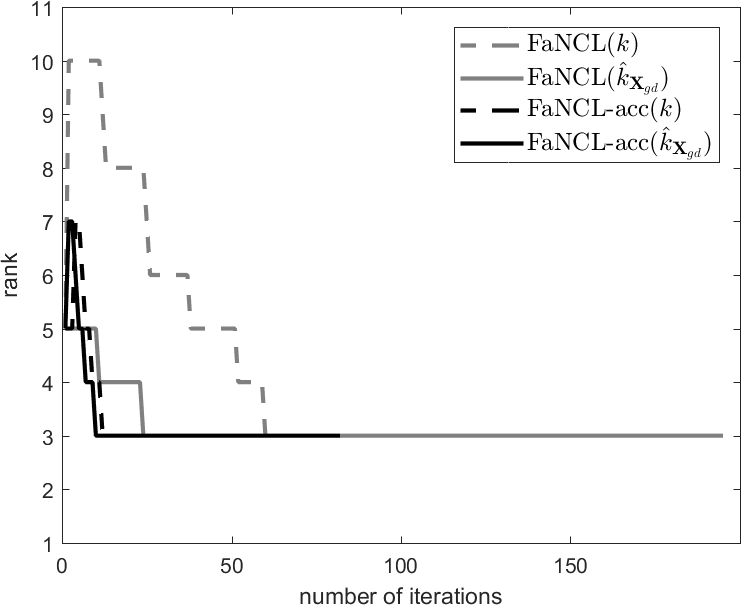}}
	\subfigure[LSP.]
	{\includegraphics[width = 0.24\textwidth]{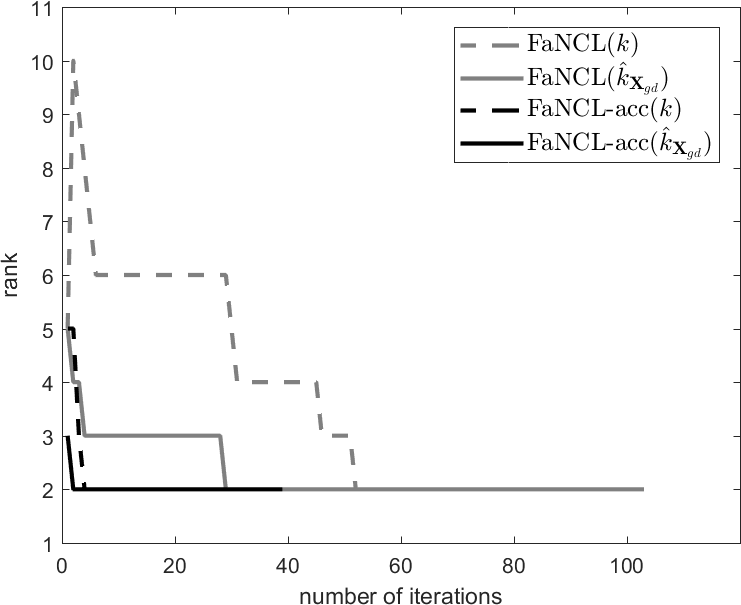}}
	
	\vspace{-10px}
	\caption{$k$ and $\hat{k}_{\X_{\text{gd}}}$ 
		vs the number of iterations 
		on the 
		\textit{MovieLens-100K} data set.
	The plot of TNN is similar and thus not shown.}
	\label{fig:rankmatcomp2}
\end{figure}

Figure~\ref{fig:rankmatcomp2} shows the ranks $k$ and $\hat{k}_{\X_{\text{gd}}}$ (as defined in Proposition~\ref{pr:apprGSVT}) vs the number of iterations on the \textit{MovieLens-100K} data set.  Recall that the data matrix is of size $943 \times 1682$.  Again, $k \ge \hat{k}_{\X_{\text{gd}}}$ and the ranks of the iterates obtained by both algorithms are low.

\noindent
\textbf{Netflix and Yahoo}:
Next, we perform experiments on two very large recommendation
data sets,
\textit{Netflix}
and \textit{Yahoo}
(Table~\ref{tab:recSys}).
We randomly use $50\%$ 
of the observed ratings 
for training,
$25\%$ for validation and the rest for testing.
Each experiment is repeated five times. 

\begin{figure}[ht]
\centering
\subfigure[\textit{netflix}.]
{\includegraphics[width = 0.24 \textwidth]{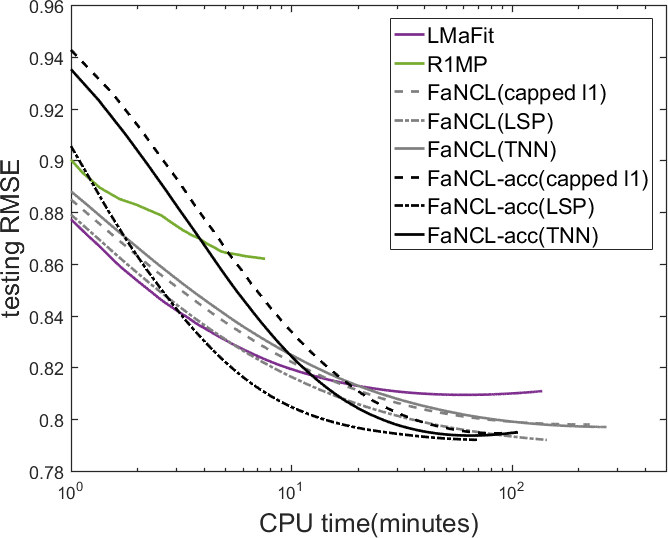}}
\subfigure[\textit{yahoo}.]
{\includegraphics[width = 0.24 \textwidth]{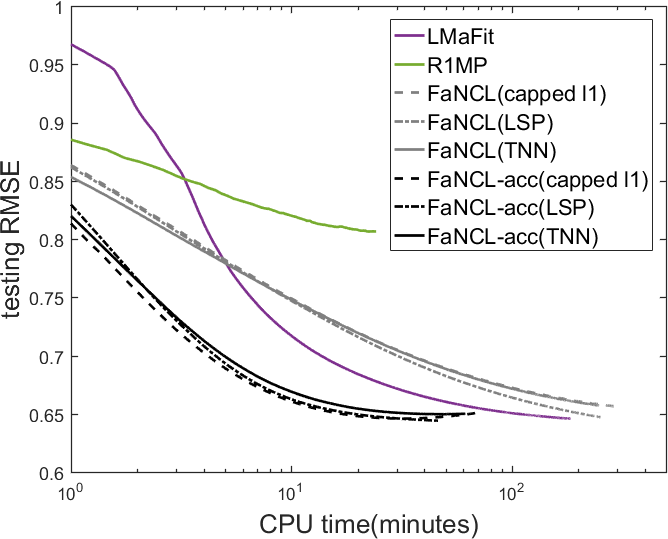}}

\vspace{-10px}
\caption{RMSE vs CPU time on the \textit{netflix} and \textit{yahoo} data sets.}
\label{fig:large:rmse}
\end{figure}

Results are shown in Table~\ref{tab:movielen}.
\textsf{APG}, \textsf{GPG} and \textsf{IRNN} cannot be run as the
data set
is large.
From Section~\ref{sec:recsys},
\textsf{AIS-Impute} has similar running time as \textsf{LMaFit} but inferior
performance,
and thus is not compared.
Again, the nonconvex regularizers converge faster, yield lower RMSE's and solutions of much
lower ranks.
Figure~\ref{fig:large:rmse} shows the RMSE vs time,
and \textsf{FaNCL-acc} is the fastest.

\begin{table*}[ht]
\centering
\caption{Results on grayscale image impainting. CPU time is in seconds.}
\vspace{-10px}

\begin{tabular}{c|c|c c c|c c c|c c c}
	\hline
	        &                     &           \multicolumn{3}{c|}{\textit{tree}}           &        \multicolumn{3}{c|}{\textit{rice}}        &            \multicolumn{3}{c}{\textit{wall}}            \\
	        &                     &           RMSE           & rank &         time         &           RMSE           & rank &      time      &           RMSE           & rank &         time          \\ \hline
	nuclear &    \textsf{APG}     &     0.433$\pm$0.001      & 180  &    308.7$\pm$71.9    &     0.224$\pm$0.002      & 148  & 242.3$\pm$27.8 &     0.222$\pm$0.001      & 165  &    311.5$\pm$31.2     \\ \cline{2-11}
	 norm   & \textsf{AIS-Impute} &     0.432$\pm$0.001      & 181  &    251.6$\pm$61.7    &     0.225$\pm$0.002      & 150  & 138.1$\pm$5.7  &     0.223$\pm$0.001      & 168  &    156.6$\pm$11.1     \\ \cline{2-11}
	        &   \textsf{active}   &     0.445$\pm$0.001      & 222  &   935.5$\pm$117.9    &     0.263$\pm$0.002      & 170  & 640.8$\pm$10.8 &     0.258$\pm$0.001      & 189  &    739.1$\pm$61.8     \\ \hline\hline
	 fixed  &   \textsf{LMaFit}   &     0.518$\pm$0.012      &  9   &     5.4$\pm$1.1      &     0.281$\pm$0.031      &  10  &  7.8$\pm$1.6   &     0.229$\pm$0.006      &  10  &      9.2$\pm$2.1      \\ \cline{2-11}
	 rank   &   \textsf{ER1MP}    &     0.473$\pm$0.001      &  19  &     1.5$\pm$0.2      &     0.295$\pm$0.002      &  22  &  2.2$\pm$0.6   &     0.269$\pm$0.003      &  20  &      1.2$\pm$0.1      \\ \hline\hline
	  LSP   &    \textsf{IRNN}    &     0.416$\pm$0.003      &  12  &    204.9$\pm$19.0    & \textbf{0.197$\pm$0.003} &  15  & 480.0$\pm$50.5 &     0.196$\pm$0.001      &  17  &     562.5$\pm$0.8     \\ \cline{2-11}
	        &    \textsf{GPG}     &     0.417$\pm$0.004      &  12  &    195.9$\pm$17.5    & \textbf{0.197$\pm$0.003} &  15  & 464.3$\pm$55.0 & \textbf{0.195$\pm$0.001} &  17  &    557.6$\pm$17.1     \\ \cline{2-11}
	        &   \textsf{FaNCL}    &     0.416$\pm$0.002      &  12  &     10.6$\pm$1.8     & \textbf{0.198$\pm$0.004} &  15  &  25.0$\pm$1.8  &     0.199$\pm$0.005      &  17  &     27.5$\pm$1.0      \\ \cline{2-11}
	        & \textsf{FaNCL-acc}  & \textbf{0.414$\pm$0.001} &  12  & \textbf{5.4$\pm$0.8} & \textbf{0.197$\pm$0.001} &  15  &  8.4$\pm$1.2   & \textbf{0.194$\pm$0.001} &  17  & \textbf{11.7$\pm$0.6} \\ \hline
\end{tabular}
\label{tab:gray}
\end{table*}

\subsubsection{Image Data Sets}
\label{sec:expgray}

\begin{figure}[ht]
\centering
\subfigure[\textit{rice}
(854$\times$960).
\label{fig:gray:rice}]
{\includegraphics[width = 0.13\textwidth]{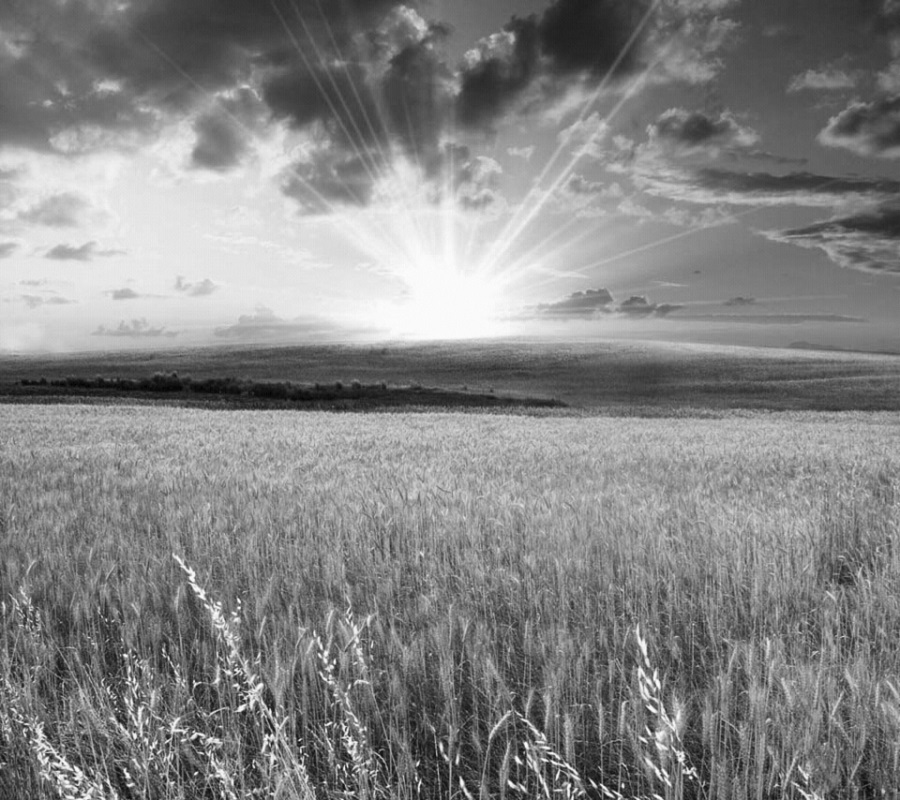}}
\quad
\subfigure[\textit{tree}.
(800$\times$800).
]
{\includegraphics[width = 0.13\textwidth]{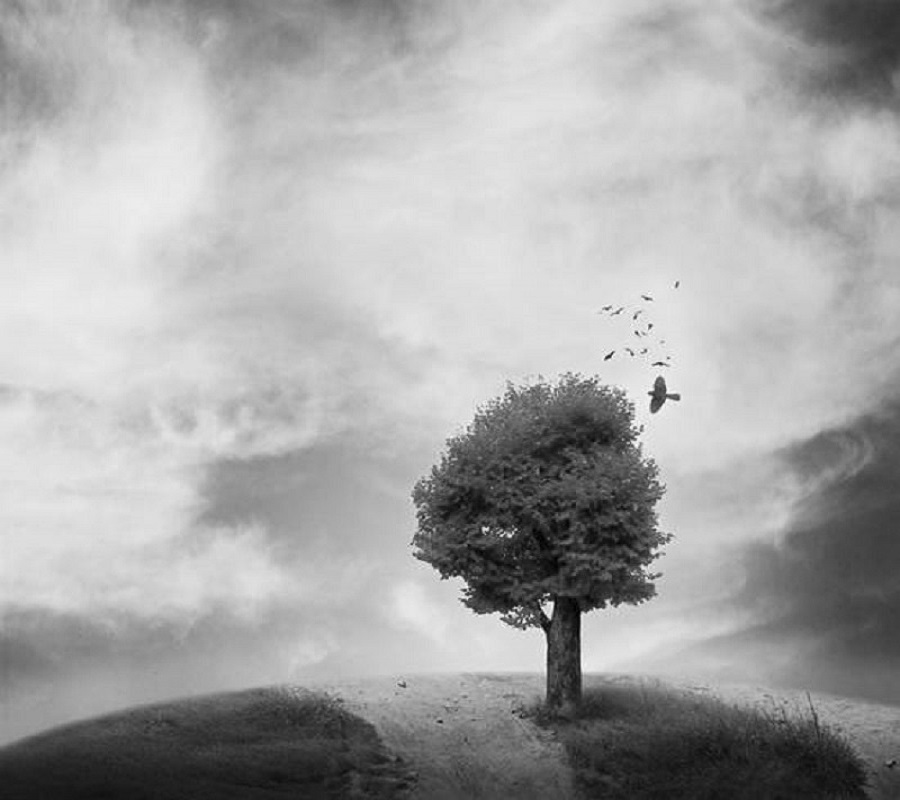}}
\quad
\subfigure[\textit{wall}
(841$\times$850).
]
{\includegraphics[width = 0.13\textwidth]{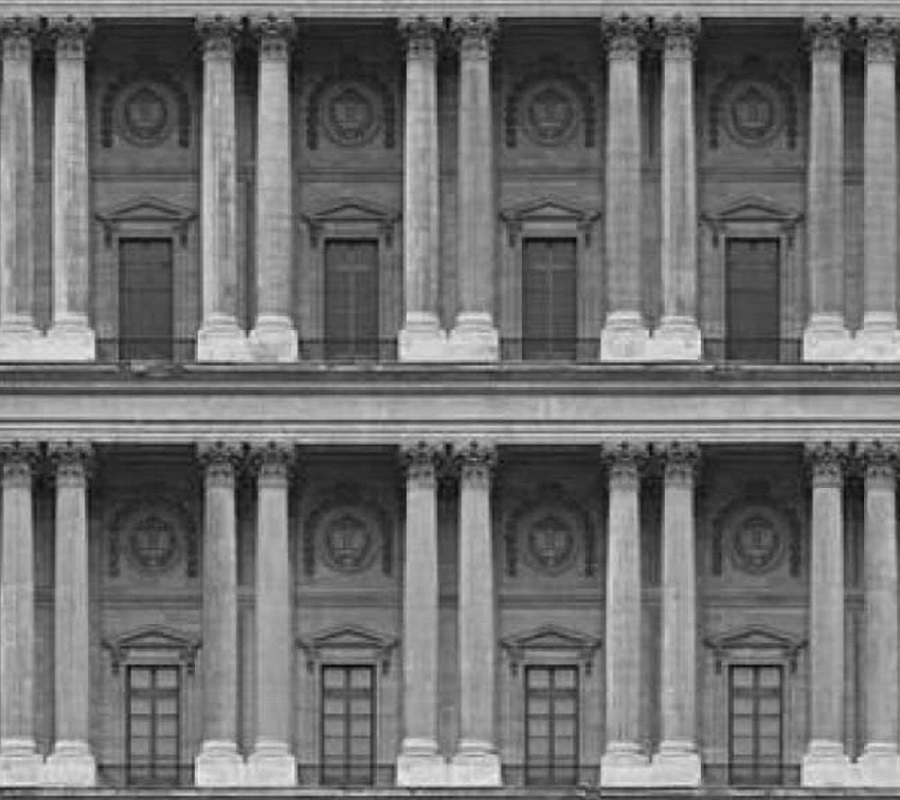}
\label{fig:gray:wall}}

\vspace{-10px}
\caption{Grayscale images used in the experiment.}
\label{fig:gray}
\end{figure}

\noindent
\textbf{Grayscale Images:}
We use the images in \cite{hu2013fast}
(Figures~\ref{fig:gray}).
The
pixels are normalized
to zero mean and unit variance.
Gaussian noise from $\mathcal{N}(0, 0.05)$ 
is then added.
In each  image,
20\% of the pixels 
are randomly sampled as observations (half for training and the half for validation).
The task is to fill in the remaining 80\% of the pixels.
The experiment is repeated five times.
The LSP regularizer is used, as 
it usually has comparable or better performance than the capped-$\ell_1$
and TNN regularizers (as can be seen from Sections~\ref{sec:matcomp:syn} and \ref{sec:recsys}).
The experiment is repeated five times.

\begin{figure}[ht]
	\centering
	\subfigure[\textit{rice}.]
	{\includegraphics[width = 0.2425\textwidth]{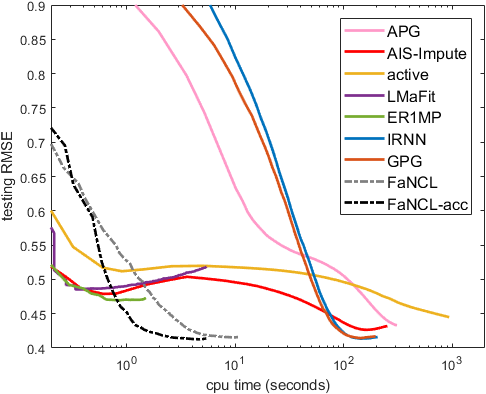}}
	\subfigure[\textit{tree}.]
	{\includegraphics[width = 0.24\textwidth]{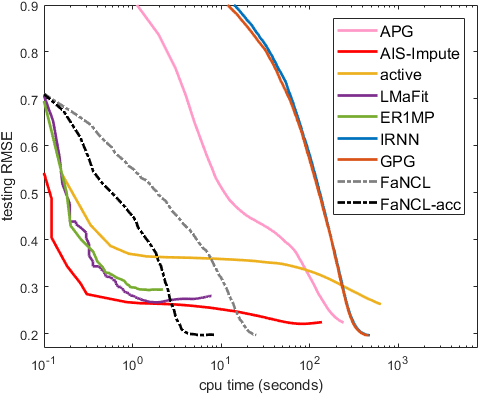}}
	
	\vspace{-10px}
	\caption{Testing RMSE vs CPU time (in seconds) on the grayscale images.
	The plot of \textit{wall} is similar and thus not shown.}
	\label{fig:rmse:image}
\end{figure}

Table~\ref{tab:gray} shows
the testing RMSE, rank obtained, and running time.
As can be seen, models based on low-rank
factorization (\textsf{LMaFit} and \textsf{ER1MP})
and nuclear norm regularization (\textsf{AIS-Impute})
have higher testing RMSE's than those using LSP regularization (\textsf{IRNN}, \textsf{GPG}, \textsf{FaNCL}, and \textsf{FaNCL-acc}).
Figure~\ref{fig:rmse:image}
shows 
convergence of the testing RMSE.
Among the LSP regularization 
algorithms,
\textsf{FaNCL-acc} is the fastest, which is then followed by
\textsf{FaNCL}, \textsf{GPG}, and \textsf{IRNN}.

\begin{figure}[ht]
\centering
\subfigure[\textit{broccoli}.
\label{fig:gray:broccoli}]
{\includegraphics[width = 0.15\textwidth]{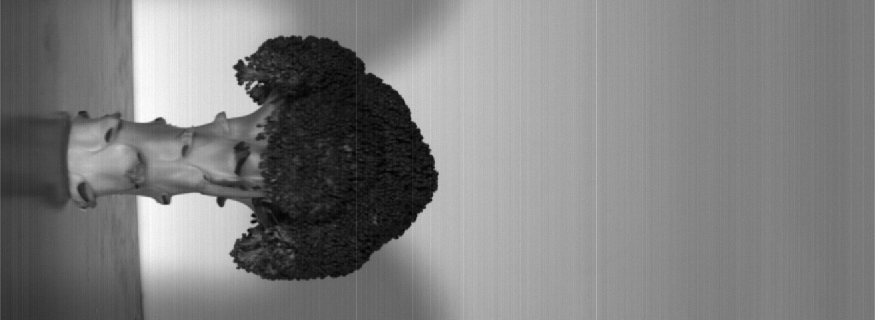}}
\quad
\subfigure[\textit{cabbage}.]
{\includegraphics[width = 0.14\textwidth]{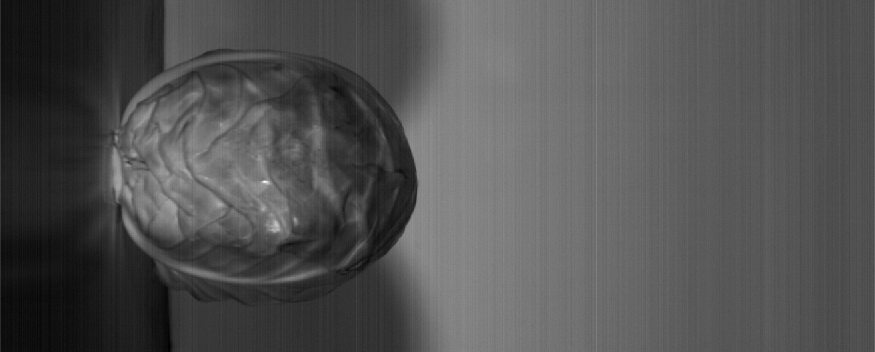}}
\quad
\subfigure[\textit{corn}.
\label{fig:gray:corn}]
{\includegraphics[width = 0.15\textwidth]{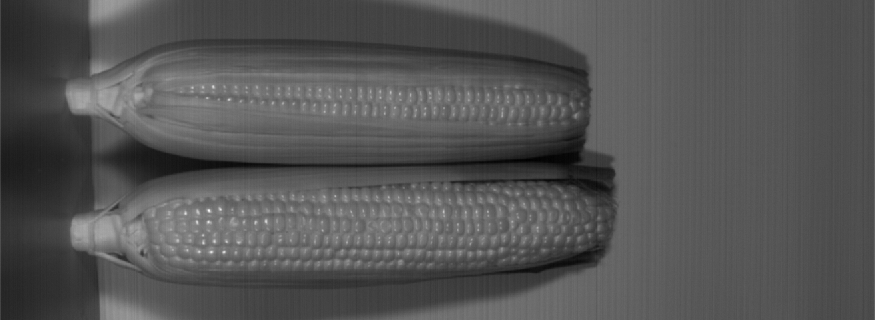}}

\vspace{-10px} 
\caption{Hyperspectral images used in the experiment, 
their sizes are 1312$\times$480$\times$49, 1312$\times$528$\times$49 and 1312$\times$480$\times$49, respectively.
One sample band of each image is shown.}
\label{fig:hyper}
\end{figure}

\noindent
\textbf{Hyperspectral Images:}
In this experiment,
hyperspectral images
(Figure~\ref{fig:hyper}) are used.
Each sample
is a 
$I_1 \times I_2 \times I_3$
tensor, where
$I_1\times I_2$ is the image size,
and $I_3$ is the number of frequencies
used to scan the object.
As in \cite{signoretto2011tensor},
we convert
this to a
$I_1 I_2 \times I_3$
matrix.
The pixels 
are normalized
to zero mean and unit variance,
and Gaussian noise from $\mathcal{N}(0, 0.05)$ is added.
1\% of the pixels are 
randomly sampled 
for training,
0.5\% for validation and 
the remaining
for testing. 
Again, we use the LSP regularizer.
The experiment is repeated five times.
As \textsf{IRNN} and \textsf{GPG} are slow (Sections~\ref{sec:matcomp:syn} and \ref{sec:recsys}), 
while \textsf{APG} and active subspace selection 
have been shown 
to be inferior to
\textsf{AIS-Impute} on the greyscale images, 
they will not be compared here.

\begin{figure}[ht]
\centering
\subfigure[\textit{broccoli}.]
{\includegraphics[width = 0.24\textwidth]{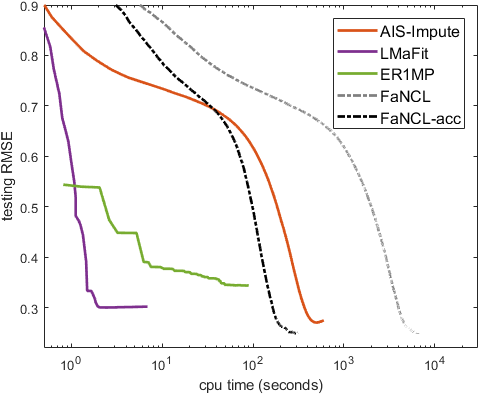}}
\subfigure[\textit{cabbage}.]
{\includegraphics[width = 0.24\textwidth]{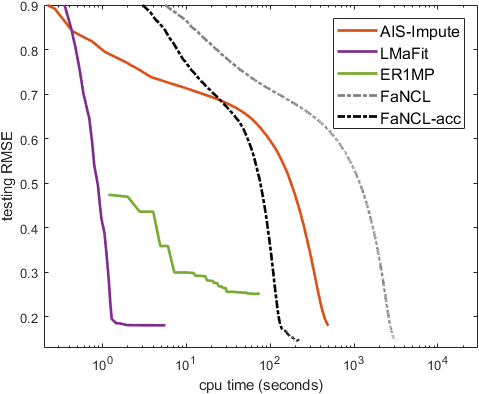}}

\vspace{-10px}
\caption{Testing RMSE vs CPU time (in seconds) on the hyperspectral images.
	The plot of \textit{corn} is similar and thus not shown.}
\label{fig:rmse:hyper}
\end{figure}

Table~\ref{tab:hyper} shows the testing RMSE,  rank obtained and running time.
As on grayscale images,
\textsf{FaNCL} and \textsf{FaCNL-acc} have lower testing RMSE's than models based on low-rank factorization 
(\textsf{LMaFit} and \textsf{ER1MP})
and nuclear norm regularization
(\textsf{AIS-Impute}). 
Figure~\ref{fig:rmse:hyper} shows convergence of
the testing RMSE.
Again,
\textsf{FaNCL-acc} is much faster than \textsf{FaNCL}.

\begin{table*}[ht]
\centering
\caption{Results on hyperspectral image impainting.
	CPU time is in seconds.
	}

\vspace{-10px}
\begin{tabular}{c|c|c c c|c c c|c c c}
	\hline
           &                     &        \multicolumn{3}{c|}{\textit{broccoli}}         &         \multicolumn{3}{c|}{\textit{cabbage}}         &             \multicolumn{3}{c}{\textit{corn}}             \\
           &                     &           RMSE           & rank & time                &           RMSE           & rank & time                &           RMSE           & rank & time                    \\ \hline
	nuclear norm & \textsf{AIS-Impute} &     0.275$\pm$0.001      & 34   & 560$\pm$90          &     0.180$\pm$0.001      & 28   & 493$\pm$46          &     0.236$\pm$0.001      & 41   & 519.1$\pm$94.7          \\ \hline\hline
 fixed rank  &   \textsf{LMaFit}   &     0.302$\pm$0.001      & 2    & 6$\pm$2             &     0.181$\pm$0.001      & 3    & 6$\pm$3             &     0.265$\pm$0.002      & 5    & 3/9$\pm$1.5             \\ \cline{2-11}
           &   \textsf{ER1MP}    &     0.344$\pm$0.002      & 21   & 88$\pm$12           &     0.252$\pm$0.003      & 41   & 75$\pm$23           &     0.299$\pm$0.004      & 37   & 55.4$\pm$7.2            \\ \hline\hline
  LSP      &   \textsf{FaNCL}    & \textbf{0.252$\pm$0.003} & 9    & 6079$\pm$1378       & \textbf{0.149$\pm$0.001} & 10   & 3243$\pm$163        & \textbf{0.204$\pm$0.004} & 15   & 4672.8$\pm$967.8        \\ \cline{2-11}
           & \textsf{FaNCL-acc}  & \textbf{0.251$\pm$0.004} & 9    & \textbf{274$\pm$77} & \textbf{0.149$\pm$0.001} & 10   & \textbf{221$\pm$12} & \textbf{0.203$\pm$0.002} & 15   & \textbf{273.8$\pm$75.4} \\ \hline
\end{tabular}
\label{tab:hyper}
\end{table*}


\subsection{Robust Principal Component Analysis}
\label{sec:expt2}


\subsubsection{Synthetic Data}
\label{sec:synrpca}

In this section, 
we first perform experiments on a synthetic data set.  The observed $m \times m$ matrix is
generated as $\mathbf{O} = \mathbf{U} \mathbf{V} + \tilde{\mathbf{S}} + \mathbf{G}$, where
elements  of
$\mathbf{U} \in \R^{m \times k}, \mathbf{V} \in \R^{k \times m}$ (with $k = 0.01m$) are 
sampled i.i.d. from $\mathcal{N}(0, 1)$,
and elements of $\mathbf{G}$ are sampled from $\mathcal{N}(0, 0.1)$.  
Matrix $\tilde{\mathbf{S}}$ is sparse, 
with $1\%$ of its elements randomly set to $5 \NM{\mathbf{U} \mathbf{V}}{\infty}$  or $- 5 \NM{\mathbf{U} \mathbf{V}}{\infty}$ with equal probabilities. 
The columns of $\mathbf{O}$ is then randomly split into training and test sets of equal size.
The standard $\ell_1$
regularizer
is 
used as the sparsity regularizer $g$ in (\ref{eq:rpca}), and
different convex/nonconvex low-rank regularizers are used as $r$.
Hyperparameters $\lambda$ and $\upsilon$ in \eqref{eq:rpca}
are tuned using the training set.  

\begin{table*}[ht]
\centering
\caption{RPCA performance on synthetic data. 
	Here, NMSE is scaled by $10^{-3}$, and  CPU time is in seconds.}
\vspace{-10px}

\begin{tabular}{cc|ccc|ccc|ccc}
	\hline
	&                    &             \multicolumn{3}{c|}{$m=500$}             &            \multicolumn{3}{c|}{$m=1000$}             &             \multicolumn{3}{c}{$m=2000$}              \\
	&                    &          NMSE          & rank &         time         &          NMSE          & rank &         time         &          NMSE          & rank &         time          \\ \hline
	nuclear norm   &    \textsf{APG}    &     4.88$\pm$0.17      &  5   &     4.3$\pm$0.2      &     3.31$\pm$0.06      &  10  &     24.5$\pm$1.0     &     2.40$\pm$0.05      &  20  &    281.2$\pm$26.7     \\ \hline\hline
	capped-$\ell_1$ &    \textsf{GPG}    & \textbf{4.51$\pm$0.16} &  5   &     8.5$\pm$2.6      & \textbf{2.93$\pm$0.07} &  10  &     42.9$\pm$6.6     & \textbf{2.16$\pm$0.05} &  20  &    614.1$\pm$64.7     \\ \cline{2-11}
	& \textsf{FaNCL-acc} & \textbf{4.51$\pm$0.16} &  5   & \textbf{0.8$\pm$0.2} & \textbf{2.93$\pm$0.07} &  10  & \textbf{2.8$\pm$0.1} & \textbf{2.16$\pm$0.05} &  20  & \textbf{24.9$\pm$2.0} \\ \hline
	LSP       &    \textsf{GPG}    & \textbf{4.51$\pm$0.16} &  5   &     8.3$\pm$2.3      & \textbf{2.93$\pm$0.07} &  10  &     42.6$\pm$5.9     & \textbf{2.16$\pm$0.05} &  20  &    638.8$\pm$72.6     \\ \cline{2-11}
	& \textsf{FaNCL-acc} & \textbf{4.51$\pm$0.16} &  5   & \textbf{0.8$\pm$0.1} & \textbf{2.93$\pm$0.07} &  10  & \textbf{2.9$\pm$0.1} & \textbf{2.16$\pm$0.05} &  20  & \textbf{26.6$\pm$4.1} \\ \hline
	TNN       &    \textsf{GPG}    & \textbf{4.51$\pm$0.16} &  5   &     8.5$\pm$2.4      & \textbf{2.93$\pm$0.07} &  10  &     43.2$\pm$5.8     & \textbf{2.16$\pm$0.05} &  20  &    640.7$\pm$59.1     \\ \cline{2-11}
	& \textsf{FaNCL-acc} & \textbf{4.51$\pm$0.16} &  5   & \textbf{0.8$\pm$0.1} & \textbf{2.93$\pm$0.07} &  10  & \textbf{2.9$\pm$0.1} & \textbf{2.16$\pm$0.05} &  20  & \textbf{26.9$\pm$2.7} \\ \hline
\end{tabular}
\label{tab:synperformance}
\end{table*}

For performance evaluation, 
we use the
(i) testing NMSE
$=\NM{(\X + \mathbf{S}) - \mathcal{P}_{\mathcal{T}}(\mathbf{UV+\tilde{S}}) }{F} / 
\NM{ 
\mathcal{P}_{\mathcal{T}}(\mathbf{UV+\tilde{S}} ) }{F}$, 
where $\mathcal{T}$ indices columns in the test set,
$\mathbf{X}$ and $\mathbf{S}$
are the recovered low-rank and sparse components, respectively;
(ii) accuracy on locating the sparse support of $\tilde{\mathbf{S}}$ (i.e.,
percentage of entries that $\tilde{S}_{ij}$ and $S_{ij}$ are nonzero or zero together);
(iii) 
the recovered rank
and (iv) CPU time.
We vary $m$ in $\{500, 1000, 2000\}$.
Each experiment is repeated five times.
Note that \textsf{IRNN}  and active subspace selection cannot be used here.
Their objectives are of the form ``smooth function plus low-rank
regularizer", but RPCA  also has a nonsmooth $\ell_1$ regularizer.
Similarly, \textsf{AIS-Impute} is only for matrix completion.
Moreover,
\textsf{FaNCL},
which has been shown to be slower than \textsf{FaNCL-acc},
will not be compared.

Results are shown in Table~\ref{tab:synperformance}.
The accuracies on locating the sparse support
are always 100\% 
for all methods, and thus  are
not shown.
Moreover, while both convex and nonconvex regularizers can perfectly recover the matrix rank and
sparse locations,
the nonconvex regularizers have lower NMSE's.
As in matrix completion, \textsf{FaNCL-acc} is much faster.
The larger
the matrix, the higher 
the speedup.


\subsubsection{Background Removal in Videos}

\begin{figure}[ht]
\centering
\subfigure[{\em bootstrap}.]
{\includegraphics[width = 0.2\columnwidth]{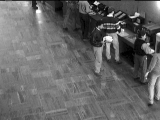}}
\subfigure[{\em campus}.]
{\includegraphics[width = 0.2\columnwidth]{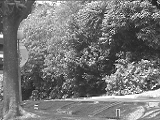}} 
\subfigure[{\em escalator}.]
{\includegraphics[width = 0.2\columnwidth]{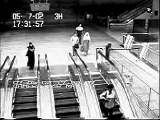}} 
\subfigure[{\em hall}.]
{\includegraphics[width = 0.2\columnwidth]{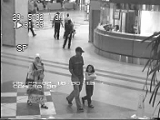}}

\vspace{-10px}
\caption{Example image frames in the videos.}
\label{fig:exampleImages}
\end{figure}

In this section, we use RPCA for background removal in videos. Four benchmark 
videos in \cite{candes2011robust,sun2013robust}
are used
(Table~\ref{tab:sumVideo}), and example frames are shown in
Figure~\ref{fig:exampleImages}.
As in \cite{candes2011robust}, the image background is considered low-rank, while the foreground moving objects contribute to the 
sparse component.

\begin{table}[ht]
\centering
\caption{Videos used in the experiment.}
\vspace{-10px}
\begin{tabular}{c | c | c | c | c}
	\hline
	                 & {\em bootstrap} & {\em campus} & {\em escalator} & {\em hall} \\ \hline
	\#pixels / frame & 19,200          & 20,480       & 20,800          & 25,344     \\ \hline
	 total \#frames  & 9,165           & 4,317        & 10,251          & 10,752     \\ \hline
\end{tabular}
\label{tab:sumVideo}
\end{table}

\begin{table*}[ht]
\centering
\caption{PSNR (in dB) and CPU time (in seconds) on the video background removal experiment. 
	The PSNRs for all the input videos are 16.47dB.}
\vspace{-10px}

\begin{tabular}{cc|cc|cc|cc|cc}
	\hline
	\multicolumn{2}{c|}{}         &    \multicolumn{2}{c|}{\textit{bootstrap}}    &      \multicolumn{2}{c|}{{\em campus}}      &     \multicolumn{2}{c|}{{\em escalator}}      &        \multicolumn{2}{c}{{\em hall}}         \\ 
	\multicolumn{2}{c|}{}         &          PSNR           &        time         &          PSNR           &       time        &          PSNR           &        time         &          PSNR           &        time         \\ \hline
	nuclear norm   &    \textsf{APG}    &     23.07$\pm$0.02      &     524$\pm$84      &     22.47$\pm$0.02      &     101$\pm$6     &     24.01$\pm$0.01      &     594$\pm$86      &     24.25$\pm$0.03      &     553$\pm$85      \\ \hline\hline
	capped-$\ell_1$ &    \textsf{GPG}    &     23.81$\pm$0.01      &    3122$\pm$284     &     23.21$\pm$0.02      &    691$\pm$43     &     24.62$\pm$0.02      &    5369$\pm$238     &     25.22$\pm$0.03      &    4841$\pm$255     \\ \cline{2-10}
	& \textsf{FaNCL-acc} &     24.05$\pm$0.01      & \textbf{193$\pm$18} &     23.24$\pm$0.02      &     53$\pm$5      & \textbf{24.68$\pm$0.02} &     242$\pm$22      &     25.22$\pm$0.03      & \textbf{150$\pm$10} \\ \hline
	LSP       &    \textsf{GPG}    &     23.93$\pm$0.03      &    1922$\pm$111     &     23.61$\pm$0.02      &    324$\pm$27     &     24.57$\pm$0.01      &    5053$\pm$369     & \textbf{25.37$\pm$0.03} &    2889$\pm$222     \\ \cline{2-10}
	& \textsf{FaNCL-acc} & \textbf{24.30$\pm$0.02} & \textbf{189$\pm$15} & \textbf{23.99$\pm$0.02} &     69$\pm$8      &     24.56$\pm$0.01      & \textbf{168$\pm$15} & \textbf{25.37$\pm$0.03} & \textbf{144$\pm$9}  \\ \hline
	TNN       &    \textsf{GPG}    &     23.85$\pm$0.03      &    1296$\pm$203     &     23.12$\pm$0.02      &    671$\pm$21     &     24.60$\pm$0.01      &    4091$\pm$195     &     25.26$\pm$0.04      &    4709$\pm$367     \\ \cline{2-10}
	& \textsf{FaNCL-acc} &     24.12$\pm$0.02      &     203$\pm$11      &     23.14$\pm$0.02      & \textbf{49$\pm$5} & \textbf{24.66$\pm$0.01} &     254$\pm$30      &     25.25$\pm$0.06      &     148$\pm$11      \\ \hline
\end{tabular}
\label{tab:videoPSNR}
\end{table*}

Given a video with $n$ image frames, each $m_1 \times m_2$ frame is first reshaped as a
$m$-dimensional column vector (where $m = m_1 m_2$), and
then all the frames are stacked together to form a $m \times n$ matrix.
The pixel values are normalized to $[0, 1]$, and
Gaussian noise from $\mathcal{N}(0, 0.15)$ is added.
The experiment is repeated five times.
For performance evaluation, we use the commonly used peak signal-to-noise ratio 
\cite{gu2016weighted}: PSNR $= - 10 \log_{10}(\frac{1}{m n} \NM{\mathbf{X} - \mathbf{O}}{F}^2)$
where 
$\mathbf{X} \in \R^{m \times n}$ is the recovered video,
and $\mathbf{O} \in \R^{m \times n}$ is the ground-truth.

\begin{figure}[ht]
	\centering
	
	\subfigure[\textit{bootstrap}.]
	{\includegraphics[width = 0.24\textwidth]{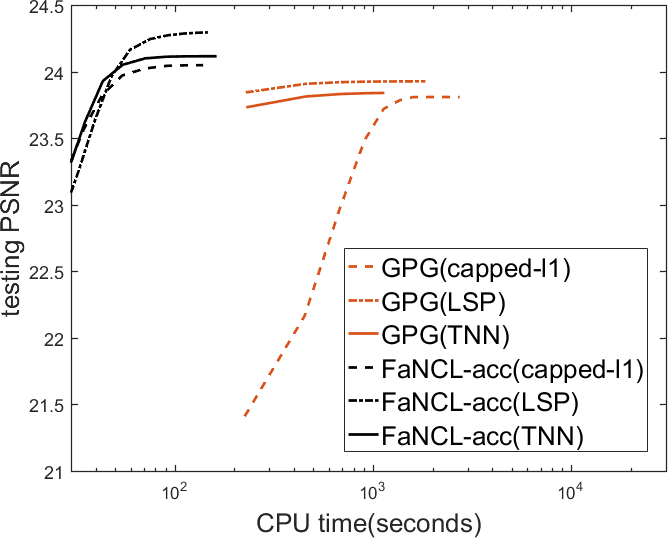}}
	\subfigure[{\em campus}.]
	{\includegraphics[width = 0.24\textwidth]{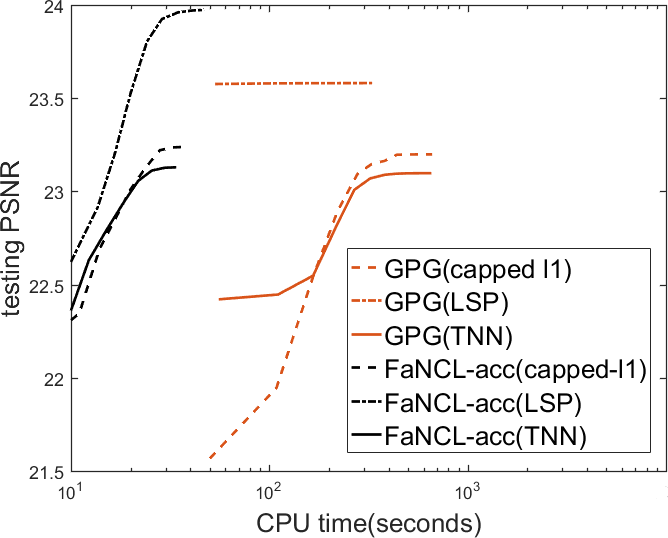}}
	
	\vspace{-10px}
	\caption{PSNR vs CPU time on the {\em bootstrap} and {\em campus} videos.}
	\label{fig:bootstrap}
\end{figure}

Results are shown in Table~\ref{tab:videoPSNR}.
As can be seen, the nonconvex regularizers lead to 
better  PSNR's than the convex nuclear norm. 
Moreover, 
\textsf{FaNCL-acc} is much faster than \textsf{GPG}.
Figure~\ref{fig:bootstrap} shows PSNR vs CPU time on the {\em bootstrap} and {\em campus} data sets.
Again, \textsf{FaNCL-acc} converges to higher PSNR much faster.
	Results on {\em hall} and {\em escalator} are similar.

\subsection{Parallel Matrix Completion}
\label{sec:expparallel}

In this section, we experiment with the proposed parallel algorithm in Section~\ref{sec:parallel}
on the \textit{Netflix} and \textit{Yahoo} data sets (Table~\ref{tab:recSys}).
We do not compare with factorization-based algorithms
\cite{yu2012scalable,recht2013parallel}, as they have 
inferior performance
(Section~\ref{sec:exptmc}).
The machine has 12 cores, and one thread is used for each core.
As suggested in \cite{yu2012scalable}, we randomly shuffle all the matrix columns and rows
before partitioning.  We use the LSP penalty (with $\theta = \sqrt{\lambda}$) and fix the
total number of iterations to 250.  The 
hyperparameters are the
same 
as in Section~\ref{sec:recsys}.
Experiments are repeated five times.

\begin{figure}[ht]
\centering
\subfigure[\textit{netflix}.]
{\includegraphics[width = 0.24\textwidth]{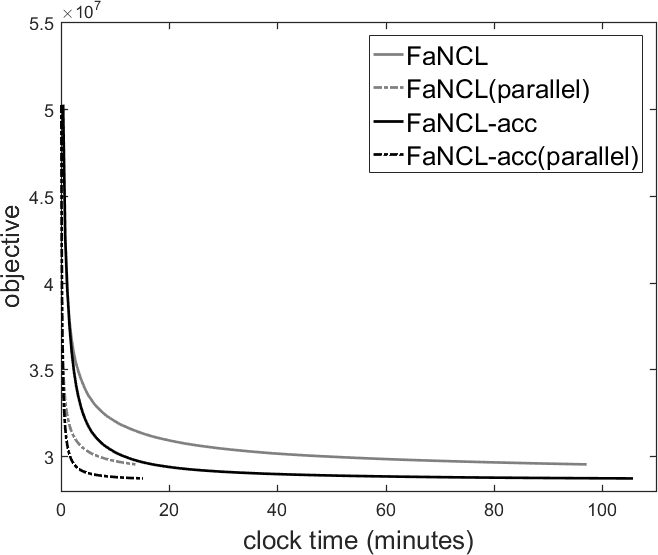}}
\subfigure[\textit{yahoo}.]
{\includegraphics[width = 0.24\textwidth]{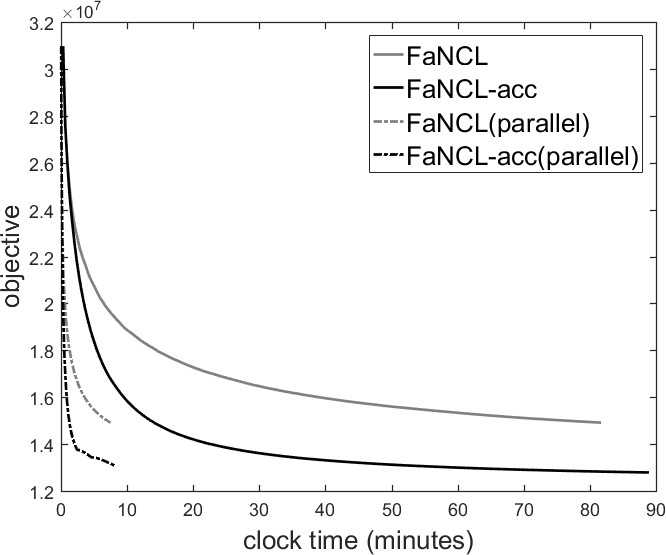}}

\vspace{-10px}
\caption{Objective value vs clock time for the sequential/parallel versions of \textsf{FaNCL} 
on the \textit{netflix} and \textit{yahoo} data sets.}
\label{fig:plobj}
\vspace{-5px}
\end{figure}

Convergence of the objective
for a typical 
run is shown in Figure~\ref{fig:plobj}.
As we have multiple threads running on a single CPU, we report the
clock time instead of CPU time.
As can be seen, the accelerated algorithms are much faster than the non-accelerated ones,
and parallelization provides further speedup.

\begin{figure}[ht]
\centering
\subfigure[Clock time per iteration. \label{fig:clock}]
{\includegraphics[width = 0.23\textwidth]{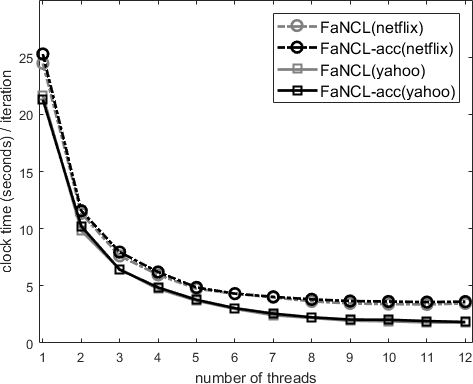}}
\subfigure[Speedup. \label{fig:speedup}]
{\includegraphics[width = 0.23\textwidth]{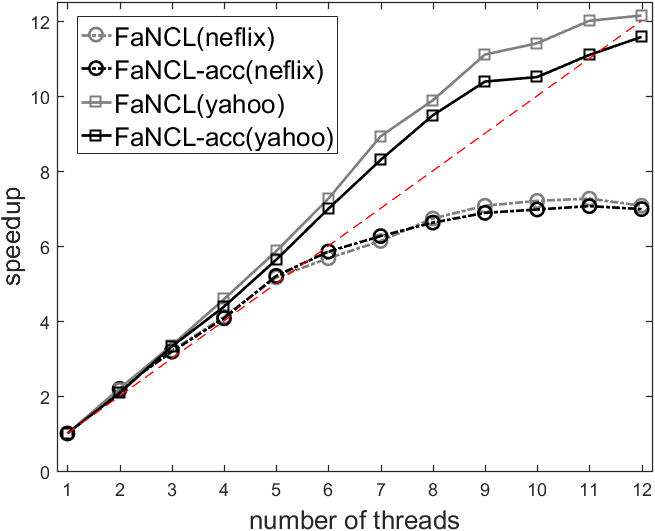}}

\vspace{-10px}
\caption{Clock time (seconds) per iteration and speedup
 vs the number of threads for parallel \textsf{FaNCL}.  
The dashed line in 
Figure~\ref{fig:speedup}
corresponds to linear speedup.}
\label{fig:paraspeedup}
\end{figure}

Figure~\ref{fig:clock} shows the 
per-iteration 
clock time 
with different numbers of threads.
As can be seen, the clock time decreases significantly with the number of threads.
Note that the curves for \textsf{FaNCL} and \textsf{FaNCL-acc} overlap.
This is because of the per-iteration time complexity of \textsf{FaNCL-acc} 
is only slightly higher than that of \textsf{FaNCL} (Section~\ref{sec:accFaNCL}).
Figure~\ref{fig:speedup} shows the speedup with different numbers of threads.
In particular, scaling is
better on \textit{yahoo}.
The observed entries in its partitioned data submatrices 
are distributed more evenly, 
which improves performance of parallel algorithms
\cite{gemulla2011large}.
Another observation  is that the speedup can be larger
than one.  As discussed in \cite{bertsekas1997parallel}, in performing multiplications
with a large sparse matrix, a significant amount of time is spent on indexing its nonzero
elements.  When the matrix is partitioned, each submatrix becomes smaller and easier to be  indexed.
Thus, the memory cache also becomes more effective.

\section{Conclusion}

In this paper, we considered the challenging problem of nonconvex low-rank matrix optimization.
The key observations are that for  the popular low-rank regularizers, 
the singular values 
obtained from the proximal operator
can be automatically thresholded, and the proximal operator can be
computed on a smaller matrix.  This allows the proximal operator to be efficiently
approximated by the power method. 
We extended the proximal algorithm in this nonconvex optimization setting with acceleration and inexact proximal step.
We further parallelized the proposed algorithm,
which scales well w.r.t. the number of threads.
Extensive experiments on matrix completion and RPCA 
show that the proposed algorithm is much faster than the state-of-the-art.
It also demonstrates that nonconvex low-rank regularizers outperform the  standard
(convex) nuclear norm regularizer.


\section*{Acknowledgment}

This research was supported in part by
the Research Grants Council of the Hong Kong Special Administrative Region
(Grant 614513),
Microsoft Research Asia 
and 4Paradigm.

\bibliographystyle{IEEEtran}
\bibliography{bib}

\end{document}